\documentclass[10pt]{article}

\usepackage{booktabs}
\usepackage{multirow}
\usepackage[table]{xcolor}
\usepackage{tcolorbox}
\usepackage{makecell}

\usepackage[ lmargin=0.8in, rmargin=0.8in, tmargin=0.8in, bmargin=1in ]{geometry}
\usepackage[utf8]{inputenc}
\usepackage{graphicx}

\usepackage{amsmath, amsfonts}
\usepackage{siunitx}

\usepackage[ruled, linesnumbered]{algorithm2e}
\SetKwComment{Comment}{$\triangleright$\ }{}

\usepackage[section]{placeins}
\usepackage{float}
\usepackage{caption}
\usepackage{subcaption}

\renewcommand{\figurename}{Fig.}

\usepackage{nameref}
\usepackage[colorlinks=false, linkbordercolor={1 0 0}, citebordercolor={1 1 0}, urlbordercolor={0 0 1}]{hyperref}

\usepackage{fancyhdr}
\usepackage[numbers]{natbib}
\usepackage{enumitem}
\definecolor{Gray}{gray}{0.25}
\definecolor{DeepGreen}{rgb}{0.0, 0.5, 0.0}

\usepackage{listings}
\usepackage{setspace}
\usepackage{wrapfig}
\usepackage{sidecap}

\makeatletter
\renewcommand{\@biblabel}[1]{\quad#1.}
\makeatother

\begin{document}

\begin{flushleft} {\Large

\textbf\newline{PrecLLM: A Privacy-Preserving Framework for Efficient Clinical Annotation Extraction from Unstructured EHRs using Small-Scale LLMs} } \newline Yixiang Qu$^\dagger$\textsuperscript{1}, Yifan Dai$^\dagger$\textsuperscript{1}, Shilin Yu$^\dagger$\textsuperscript{1}, Pradham Tanikella\textsuperscript{2,3}, 
Malvika Pillai\textsuperscript{4},
Walter Chen\textsuperscript{1},
Jialiu Xie\textsuperscript{1},
Yishan Ren\textsuperscript{1}, Duan Wang\textsuperscript{1},  Yikai Wang\textsuperscript{5},
Sid Sheth\textsuperscript{6,10}, Guanting Chen\textsuperscript{5}, Yufeng Liu\textsuperscript{7}, Travis Schrank\textsuperscript{6}, Trevor Hackman\textsuperscript{6}, Didong Li\textsuperscript{1, 5,10}, Di Wu\textsuperscript{1,4,8,9,10,*}

\bigskip \bf{1} Department of Biostatistics, University of North Carolina at Chapel Hill \\ 
\bf{2} Department of Genetics, University of North Carolina at Chapel Hill \\
\bf{3} Curriculum for Bioinformatics and Computational Biology, University of North Carolina at Chapel Hill at Chapel Hill\\
\bf{4} Carolina Health Informatics Program, University of North Carolina at Chapel Hill \\
\bf{5} Department of Statistics and Operations Research, University of North Carolina at Chapel Hill\\ 
\bf{6} Department of Otolaryngology/Head and Neck Surgery, University of North Carolina at Chapel Hill\\ \bf{7} Department of Statistics, University of Michigan\\ 
\bf{8} Department of Biomedical Sciences, Adams School of Dentistry, University of North Carolina at Chapel Hill\\ 
\bf{9} Computational Medicine Program, University of North Carolina at Chapel Hill at Chapel Hill\\
\bf{10} Lineberger Comprehensive Cancer Center, University of North Carolina at Chapel Hill\\ 
\bigskip * dwu@unc.edu

\end{flushleft}

\section*{Abstract}
Large Language Models (LLMs) have demonstrated remarkable proficiency in automated text annotation within natural language processing. However, their deployment in clinical settings is severely constrained by strict privacy regulations and the prohibitive computational cost of processing voluminous unstructured Electronic Health Records (EHRs).  Unstructured EHR typically include crucial information for clinical decisions in precision medicine, such as in cancer. To overcome this bottleneck and enable scalable annotation extraction from clinical notes, we developed a compact LLM framework featuring an novel EHR-specific resource-efficient PREproCessing technique that can be adopted in existing LLM procedures (PrecLLM). This technique is particularly useful for the smaller LLMs which are often more accuracy-challenged. PrecLLM has been optimized for local deployment in computational environments with stringent privacy requirements and restricted access to high-performance GPUs. Two alternative simple yet powerful procedures have been provided in the preprocessing step includes both regular expressions (regex) and Retrieval-Augmented Generation (RAG) to extract and highlight key information from unstructured clinical notes. Pre-filtering long and unstructured texts enhanced the performance of smaller LLMs on EHR-related tasks. Evaluation was performed on two distinct cohorts: a locally curated private EHR dataset from the EPIC system for a Head and Neck Cancer (HNC) cohort, and the publicly available EHR dataset (MIMIC-IV). Using MIMIC-IV, we further compared PrecLLM against fine-tuned LLMs. Results demonstrated that PrecLLM substantially enhanced the performance of the original smaller LLMs in terms of sensitivity, specificity, and F1 scores, making it well-suited for privacy-sensitive and resource-constrained applications. This study offers optimized LLM performance for local, secure, and efficient healthcare applications, and provides practical guidance for clinical LLM deployment while addressing challenges related to privacy, computational feasibility, and clinical applicability. The full implementation of the PrecLLM pipeline is available at \href{https://github.com/renlyly/LLM_ClinicalNote}{https://github.com/renlyly/LLM\_ClinicalNote}.

\section{Introduction}

Electronic Health Records (EHRs) are crucial for prognosis, diagnosis, and treatment decisions in modern healthcare as they contain comprehensive patient data \cite{desRoches2008electronic}, including both structured data (e.g., demographics, medications, lab results, genomics) and unstructured data (e.g., clinical notes). The combination of both structured and unstructured data enhances personalized clinical decisions, drug discovery, and informative policy-making.  The improved annotation of clinical phenotypes will enhance prediction of outcomes, such as mental disease diagnosis or cancer treatment effects, and further improve precision care of patients. Effectively harnessing unstructured data, despite the inherent challenges, unlocks significant information that is often unrecorded in structured data, only partially available for many subjects, or more accurate than their structured counterparts. For instance, despite advances in EHR systems, the lack of precise disease phenotyping in structured data remains a major challenge for their full use. While the International Classification of Diseases (ICD) codes are recorded in most EHRs, they often fail to capture the details of a patient's condition \cite{harris2020icd}. In contrast, clinical notes can more accurately reflect disease progression, but their unstructured nature makes it challenging to extract relevant information. 

To manage the complexity of unstructured EHR data, early computational phenotyping strategies relied on keyword searching. These methods typically extracted disease, medication, or lab-related keywords using lexical matching algorithms, subsequently deriving disease phenotypes based on rules designed by domain experts \cite{zeng2019phenotyping}. Rule-based systems leveraging resources such as the Unified Medical Language System \cite{Bodenreider2004rule} were utilized for various phenotypes including diabetes mellitus \cite{ware2009rule} and adverse drug events \cite{li2014rule}. However, these designed rules often fall short in inferring complex clinical information from notes. To address this limitation, some studies utilized statistical learning methods to predict disease phenotypes after engineering features from the searched keywords and structured data \cite{pineda2015comparison, teixeira2017evaluating, zhang2019high}. While these methods improved upon manually designed rules, they typically ignored the context of the searched keywords within the clinical notes, potentially leading to misinterpretation.

Transformer-based models, such as Bidirectional Encoder Representations from Transformers (BERT) \cite{devlin2019bert}, have gained substantial popularity and importance in recent years due to their remarkable ability to understand and generate human-like text. These models leverage vast amounts of data and complex neural network architectures to deliver insights and automation across various fields. A popular approach to effectively learn disease phenotypes from clinical notes is to fine-tune pre-trained models, such as clinical BERT \cite{alsentzer2019clinicalbert}, using annotated clinical documents. However, these fine-tuned models have primarily been developed for clinical text mining and Named Entity Recognition (NER) \cite{peng2019transfer, lee2020biobert, gu2021domain}, and are less suited for disease phenotyping, potentially due to the lack of publicly available annotated data. Disease phenotyping often involves training the models on a representative subset of cohort records, manually annotated by a team of experts, to subsequently automate annotation for the larger remaining dataset \cite{yang2023machine}. However, manual annotation of these notes is labor-intensive, prone to errors \cite{Sylolypavan2023annotation}, and requires highly skilled healthcare professionals. Zhang et al fine-tuned BERT to extract comprehensive collections of breast cancer-related phenotypes using clinical notes from a Chinese hospital \cite{zhang2019extracting}. Zhou et al adopted a similar approach to develop a cancer domain-specific BERT model (CancerBERT) for breast cancer-related phenotype extraction from the University of Minnesota Clinical Data Repository \cite{zhou2022cancerbert} . Hartman et al fine-tuned BERT and BART models to automate the generation of discharge summaries for neurology patients \cite{hartman2023method}. Notably, all these applications require highly skilled clinicians to annotate notes from large-scale datasets, which can be labor-intensive and time-consuming.

Large Language Models (LLMs), such as GPT \cite{radford2018gpt}, {Gemma} \cite{team2024gemma}, and {LLaMA} \cite{touvron2023llama}, present a promising approach to automate processes like disease phenotype extraction, significantly reducing the associated labor and time. This is largely because these models can often directly address such tasks or answer relevant questions without requiring extensive, task-specific expert annotation. Several studies have employed LLMs for various clinical tasks \cite{li2024scoping}, often leveraging their ability to understand contextual language and generate human-like text across a wide range of Natural Language Processing (NLP) tasks \cite{fan2024bibliometric, fan2024nphardeval}. Some studies employ these LLMs directly – for instance, GPT-3.5 has been used to extract social determinants and family history \cite{bhate2023zero}, while both GPT-3.5 and GPT-4 have generated draft responses for patient messages \cite{garcia2024artificial}. Another common strategy involves fine-tuning models for specific clinical contexts; \cite{Jiang2023llmnature}, for example, proposed NYUTron, an LLM pretrained and fine-tuned on hospital notes for comprehensive patient profiling. \cite{lopez2025clinical} incorporated a more complicated method called Clinical Entity Augmented Retrieval (CLEAR), which applies NER, then entity filtering and augmentation using ontologies or LLMs, followed by chunk retrieval for subsequent LLM analysis. However, the deployment of large-scale LLMs typically requires cloud-based infrastructure. The nature of medical records demands strict adherence to privacy regulations \cite{fernandez2013security}, which limits the use of cloud-based LLMs such as ChatGPT. Many of these studies utilize cloud-based LLM services such as the OpenAI API rather than locally deployed models, which could result in privacy issues with more sensitive EHRs. In addition, these models demand significant computational power, particularly high-performance GPUs with large amounts of Video Random Access Memory (VRAM), which may not be accessible to many labs \cite{musser2023cost}.

\begin{figure*}[ht] 
\centering 
\includegraphics[width=0.8\textwidth]{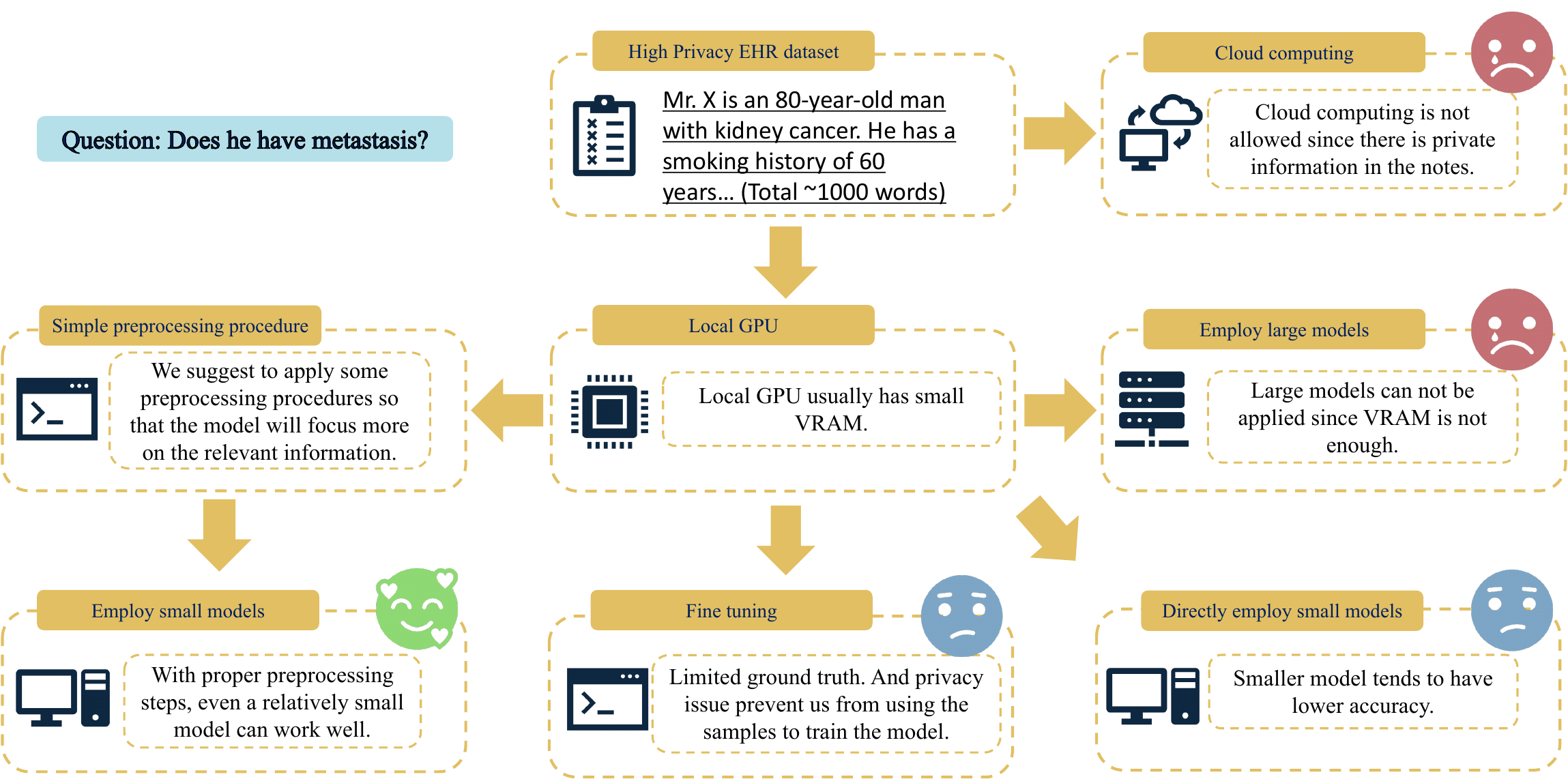} 
\caption{Schematic diagram illustrating the challenges of clinical annotation in resource-constrained environments and the rationale behind PrecLLM. 
\textbf{Constraints (Right):} Processing high-privacy EHR datasets prohibits the use of \textbf{cloud computing} due to data leakage risks. 
\textbf{Hardware Limitations (Center):} Local deployment is constrained by the limited VRAM of standard GPUs, which makes running \textbf{large models} infeasible. 
\textbf{Alternative Pitfalls:} \textbf{Fine-tuning} (bottom center) is often restricted by the scarcity of annotated ground truth and privacy policies preventing training on patient data, while \textbf{directly employing small models} (bottom right) on raw text typically suffers from low accuracy. 
\textbf{Proposed Solution (Left):} PrecLLM addresses these bottlenecks by introducing a \textbf{simple preprocessing procedure} that filters noise, enabling \textbf{small-scale models} to achieve high accuracy locally without requiring extensive computational resources or labeled training data.} 
\label{fig:diagram} 
\end{figure*}

To address these challenges, we propose PrecLLM (Fig.~\ref{fig:diagram}) to enhance the performance of small LLMs by including a novel preprocessing step, as incorporating pre-filtering. We focus on small, more accessible, locally-deployed LLMs such as \texttt{Gemma-7B} \cite{team2024gemma} and \texttt{LLaMA-7B} \cite{touvron2023llama}. These models offer the advantage of reduced computational burden, making them suitable for local private usage, yet they suffer from limited context windows~\cite{liu2024longcontext, li2024longcontext}. Feeding raw, lengthy clinical histories into these models often leads to ``lost-in-the-middle'' 
phenomena, where key diagnostic signals are diluted by irrelevant noise, resulting in hallucinations or missed diagnoses. To make it feasible for these LLMs to work within their context window constraints, we introduce a well-designed pre-filtering step. Only the most annotation-relevant note sections are retained to reduce the input text length and capture key information effectively in the LLM. This is achieved through Regular Expressions (regex), which identify predefined relevant keywords, and Retrieval-Augmented Generation (RAG), which retrieves sentences based on semantic similarity when prior knowledge is limited (detailed model and implementation specifications are provided in Supplementary Note~\ref{Sec:implementation}). We note that while these are two examples to illustrate the power of preprocessing in our framework, other NLP methods can be easily integrated.

Besides being efficient, accurate, and privacy-preserving, PrecLLM also addresses the fact that EHR data per patient, by nature, have typically been collected in a longitudinal manner at many time points through numerous visits (hospital admissions in MIMIC-IV). To reflect this, we also systematically annotate clinical variables from unstructured data at three levels, per record, per visit, and per subject. The longitudinal per-visit annotation can be vital for real-time clinical decisions and studies of dynamic treatments.

To evaluate the effectiveness of PrecLLM, we apply it to extract disease phenotypes from clinical notes using both zero-shot and few-shot prompting paradigms, i.e., we vary the number of in-prompt examples provided to the LLM, comparing their performance with and without preprocessing in several popular recent small LLMs, including a foundation model (e.g., \texttt{Gemma-7B-it}), domain-specific LLMs in biomedical fields (e.g., \texttt{LLaMA-2-7B-Chat-Med} and \texttt{Bio-Medical-LLaMA-3-8B}), and also a relatively large LLM (\texttt{Meta-Llama-3-70B-Instruct}). To ensure a comprehensive evaluation spectrum, we fine-tune the existing LLMs to compare with the typical pipeline. We use two real-life EHR datasets for evaluation. The advantages of our preprocessing method PrecLLM are demonstrated on two datasets: a private dataset comprising 7,284 head-and-neck cancer (HNC) patients and the publicly available Medical Information Mart for Intensive Care MIMIC-IV dataset \cite{Johnson2023MIMICIV}. In this paper, we use ``metastasis'', critical for cancer patients, ``insulin use'', and ``hypertension'' as the target phenotype annotations to be extracted from both datasets for evaluating the accuracy of PrecLLM extraction. These three variables are chosen because they all have existing corresponding ICD codes that are considered the ground truth for evaluation. Importantly, beyond cancer or hypertension cohorts, our framework can be easily extended to other medical conditions and terminologies, making it broadly applicable across various healthcare domains. In summary, the contributions of our work are as follows:

\begin{enumerate} \item \textbf{Investigation of LLM Performance for EHR Data}: We systematically investigate the performance of relatively small LLMs for EHR datasets, identify the challenges they face, and propose solutions to enhance their effectiveness.

\item \textbf{Introduction of a Preprocessing Framework}: We propose a preprocessing framework to filter and highlight key information in long, unstructured clinical notes before feeding into LLMs, enabling smaller LLMs to make more accurate predictions. The goal of precLLM is to improve LLM feasibility and accuracy. By leveraging simple yet effective techniques like regex-based filtering and RAG, we demonstrate significant performance improvements in resource-constrained settings. {To the best of our knowledge, no previous work has highlighted the effectiveness of simple regex-based approaches in this clinical context.} \item \textbf{Evaluation Across Multiple Learning Paradigms}: Our study includes extensive experiments that compare the effectiveness of zero-shot and few-shot learning paradigms, both with and without the application of preprocessing, on EHR data. Different LLMs are also evaluated to ensure robustness. We also compare the results of our framework with fine-tuned LLMs. {Experimental results indicate that, in scenarios with limited annotated labels, preprocessing proves to be significantly more efficient and cost-effective than fine-tuning.} This analysis highlights the considerable influence of preprocessing on model performance, enabling researchers with constrained GPU resources to maintain high accuracy while adhering to strict privacy standards. \end{enumerate}

\section{Results} 
\subsection{The PrecLLM Framework}

\begin{figure*}[!ht] 
\centering 
\includegraphics[width=1.05\textwidth]{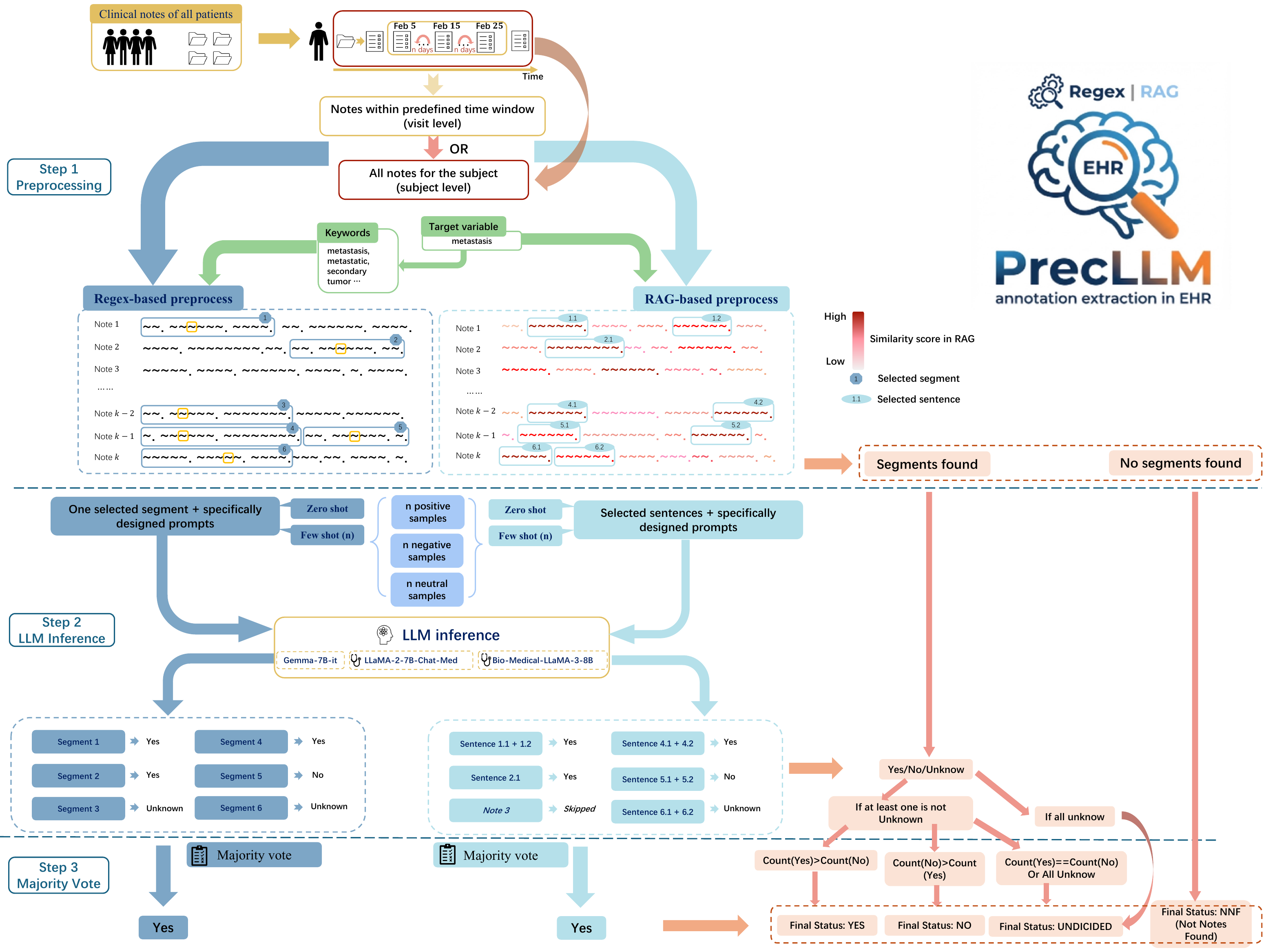} 
\caption{Flowchart of the PrecLLM framework illustrating the preprocessing methods and the decision logic for phenotype annotation. 
\textbf{Step 1 (Preprocessing):} Two filtering methods are employed to extract key information. The \textit{regex-based procedure} identifies keywords and extracts a three-sentence segment (the keyword-bearing sentence plus its immediate neighbors). The \textit{RAG-based procedure} calculates similarity scores (indicated by color depth) and selects $p$ sentences ($0 \le p \le s$) that rank in the top $s$ and exceed a predefined threshold. {If no segments are found during this step, the process terminates immediately with a status of ``No Notes Found'' (NNF).} 
\textbf{Step 2 (LLM Inference):} Successfully extracted segments are individually classified by the LLM using zero-shot or few-shot prompts into ``Yes'', ``No'', or ``Unknown''. 
\textbf{Step 3 (Majority Vote):} For subjects with valid segments, the final status is determined via the logic tree: a majority of ``Yes'' votes yields \textbf{YES}; a majority of ``No'' votes yields \textbf{NO}; whereas a tie or a set consisting entirely of ``Unknown'' outputs results in \textbf{UNDECIDED}. Throughout the diagram, an octagon represents a selected segment, while an ellipse represents a selected sentence.} 
\label{fig:detail} 
\end{figure*}

PrecLLM is a novel LLM tool allowing efficiently reduce the input volume of unstructured EHR text while preserving the accuracy and precision of annotating a target variable related to clinics.    
The PrecLLM pipeline has three steps (Fig.~\ref{fig:detail}) with a focus on preprocessing to reduce input text volume.
Here, we define the clinical variable to be annotated as the ``target variable''.
To achieve reducing the input to small LLM, filtering procedure of text was developed as the first step, relying on the assumptions that the nuances of one clinical variable to be annotated from the unstructured EHR notes may be articulated across multiple sentences and concentrated near to the Semantically Similar Terms (SST) of this clinical variable. Consequently, for regex-based keyword approaches, we extracted the sentence containing the keyword along with its immediately preceding and succeeding sentences. 
 Two points need to be addressed in the first step.
One point is to locate sentences containing SSTs. In precLLM, we provide two options of NLP methods to locate these sentences, Regular Expressions (regex) and Semantic Retrieval (part of the RAG framework). Regex and RAG both run fast, require relatively
 small computation power and are popular to be used for identifying relevant information. 
 PrecLLM offers two options to accommodate different levels of prior domain knowledge regarding the target variables. While we focus on these two methods, the framework is adaptable to other NLP techniques. The other point is to include sufficient context around the SSTs for annotation. In the regex-based approach, we retained the sentence containing the keyword plus its immediately preceding and following sentences. In addition to subject-level annotation, time-sensitive visit-level annotation is mostly needed because subjects typically had multiple visits across time, and visit timing is important for clinical decision. For the visit level, we used 10-20 days of notes before and after the visit (20-40 days total) to allow diagnostic latency and reduce missed lagged diagnoses (Step 1 in Fig.~\ref{fig:detail}). To show the performance of PrecLLM in real-world EHR data, we extracted three clinical variables: ``metastasis'' (from a local HNC cohort of over 7,000 subjects and the MIMIC-IV cohort), ``insulin use'' (from MIMIC-IV), and ``hypertension'' (from MIMIC-IV). These variables were selected because their corresponding ICD codes were available in structured EHR data and can be served as the ground truth for evaluation. The evaluation results suggest PrecLLM can be easily applied to other clinical variables and cohorts. The three steps are summarized as follows.

In the first step ``pre-filtering'', two options of NLP methods, regex and Semantic Retrieval (part of the RAG framework) are provided to search for and locate SSTs, so that sentences in text will be filtered and the input volume to LLM are to be reduced. This preprocessing step ensured that only the text segments surrounding these SSTs were retained for the subsequent annotation extraction. For details, given a subject with $k$ clinical notes within a defined time period, we segmented each note into individual sentences. Specifically, the definition of this time period was contingent on the level of analysis: for \textbf{visit-level} annotation, we defined a temporal window extending 10 to 20 days prior to and following the index visit; conversely, for \textbf{subject-level} annotation, the period encompassed the patient's entire longitudinal history. To utilize regex, a set of keywords was efficiently derived using clinicians' domain knowledge or modern language tools such as GPT-4. Keywords were defined by their semantic similarity to the target variable, encompassing attributes, name variations, and related concepts. For instance, to extract annotations for metastasis, a set of keywords including ``metastasis'' and ``metastatic'' was generated. Subsequently, we organized these keywords into a single regex search command by employing two complementary matching strategies: defining flexible patterns for structural variations (e.g., ``insulin-dependent'' with or without a hyphen), and bundling distinct synonyms into a comprehensive list. The clinicians' expertise could be helpful here to ensure the patterns and lists. The full list of ``keywords'' used in this paper for the three target variables was in the Supplementary Note~\ref{Sec:keywords}. With the keyword list obtained above, regex was used to extract the relevant sentences from the unstructured notes. Crucially, because EHR notes could be complex and might articulate the nuances of a target clinical variable across multiple sentences, we recommended extracting not only the sentence containing the keywords but also the immediately preceding and succeeding sentences. This set of three sentences was defined as one segment. This unified regex-based strategy was more rapid and comprehensive than searching for each keyword individually. Semantic Retrieval, a component of the RAG framework, served as an alternative to regex in PrecLLM for identifying informative sentences. RAG was particularly useful when domain knowledge was limited or when defining a comprehensive keyword list was difficult (e.g., when the target concept was complex). It utilized information retrieval to locate sentences semantically similar to the target variable based on cosine similarities \cite{fan2024survey}. In the RAG-based preprocessing step, sentences from each clinical note underwent a semantic similarity assessment against the target variable. From each note, we identified the top $s$ most similar sentences and selected a subset of $p$ sentences ($0 \le p \le s$) that exceeded a predefined similarity threshold. We, therefore, obtained the segments of sentences relevant to the target variable from the EHR note. Unlike regex, which segments text based on exact matches and their immediate context, RAG retrieved specific high-similarity sentences. This process could be repeated for multiple notes across subjects. If no relevant clinical notes or text segments were found after filtering, no LLM classification was possible and the final PrecLLM annotation was designated as ``NNF''.

In the second step ``annotating segments'', small-scale LLMs are applied only to the filtered segmented texts to extract the segment-level of clinical annotations. For each processed segment, the filtered text from Step 1, including the contextual information, served as the input to a small LLM, using specifically designed prompts to infer whether the visit or subject had the target clinical condition. The LLM then output a classification result for the segment as one of the three {``Yes''}, {``No''}, or {``Unknown''}. The local LLMs employed in this study included the general-purpose \texttt{Gemma-7B-it} (Version 1) and medical-specific models like \texttt{LLaMA-2-7B-Chat-Med} and \texttt{Bio-Medical-LLaMA-3-8B} (detailed specifications in Supplementary Note~\ref{Sec:implementation}). To assess PrecLLM performance on scale, the framework was also evaluated on the substantially larger \texttt{Meta-Llama-3-70B-Instruct}. In practice, a single local LLM was sufficient for our experiments. We later validated the preprocessing step across all of these models. The ``Unknown'' classifications from LLM generally manifested under two conditions: (1) particularly in RAG-based processing, the $p$ sentences might be selected even if a clinical note did not explicitly reference the target phenotype (these sentences might possess low absolute similarity scores, although above the predefined threshold), a situation less frequently encountered with regex processing due to its prerequisite of exact keyword matching; or (2) in both methods, a note might allude to the phenotype but lack definitive confirmatory or negating evidence (e.g., when diagnostic outcomes were pending).

In the third step ``summarizing via majority vote'', a summarized  clinical annotation will be output using majority vote across segments at the subject, visit or both levels. The output level depends on the input clinical notes. This is to obtain an overall status of the target variable for the subject across all time points, or within the predefined time period. Here, a majority vote (see Section~\ref{sec:majority_vote}) was applied to the extracted classifications from all processed segments for that subject after excluding the ``Unknown''. This decision was principally motivated by the observation that ``Unknown'' output typically offered limited substantive information when there were ``Yes'' or ``No'' classification results. The explicit omission of these ``Unknown'' segments or clinical notes ensured they did not influence the majority vote. This strategy was intended to preserve the integrity of the classification by concentrating the decision-making process on more conclusive data points, thereby aiming for a more reliable overall outcome. Consequently, on one hand, depending on the ratio between ``Yes'' and ``No'' in the majority vote, the summarized output in Step 3 included ``Yes'', ``No'', and ``Undecided'' (when there were equal counts of ``Yes'' and ``No''). On the other hand, if the classification results from all segments were ``Unknown'', the output of Step 3 would be ``Undecided''. One salient distinction between regex and RAG in preprocessing, which warrants careful consideration, pertains to their impact on the input to the majority vote. With regex-based preprocessing, a single clinical note containing multiple, distinct mentions of the target variable could contribute several segments and thereby multiple votes to the subject-level overall annotation. Conversely, RAG contributed one vote per note, based on its $p$ semantically similar sentences, irrespective of the frequency of explicit keyword occurrences within that note. This influenced the final results.

Overall, PrecLLM provides a preprocessing step embedded in a three-step pipeline to be used at large scale on unstructured clinical notes from EHRs for efficient and accurate annotation extraction while preserving data privacy locally. In addition, this study provided a comprehensive evaluation in two real-world EHR datasets to benchmark performance and guide the choice of parameter options in the LLM application. In the output of the PrecLLM framework for each extraction task, one of four outcomes ``Yes'', ``No'', ``Undecided'', or ``No Notes Found (NNF)'' (defined in Step 1) is assigned.

\subsection{Overview of Evaluation Procedure}

\begin{figure*}[ht!] \centering \includegraphics[width=0.7\textwidth]{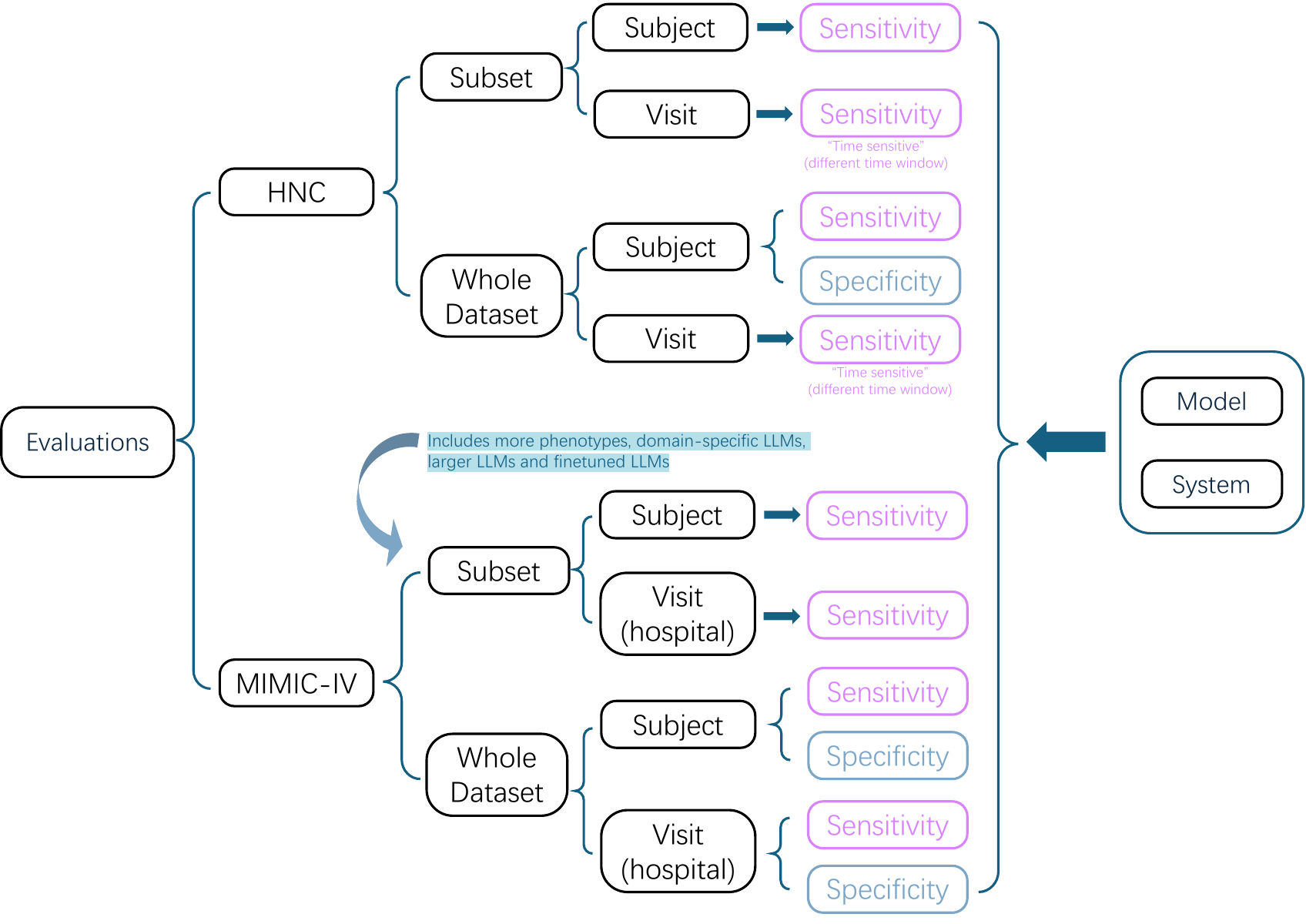} 

\caption{Hierarchical overview of the evaluation protocols for the HNC and MIMIC-IV datasets. 
\textbf{Dataset Stratification:} The evaluation is divided into private HNC and public MIMIC-IV cohorts, further split into subsets (for sensitivity analysis across multiple models) and whole datasets (for comprehensive performance assessment). 
\textbf{Evaluation Granularity:} Assessments are conducted at two levels: the patient level (\textit{Subject}) and the encounter level (\textit{Visit}, defined by time windows in HNC and hospital admissions in MIMIC-IV). 
\textbf{Metric Availability:} Sensitivity is calculated across all branches (pink boxes), whereas Specificity (blue boxes) is exclusively reported for the whole datasets due to the enrichment of positive cases in the subsets. 
\textbf{Comparison Scope:} As indicated by the arrow, the MIMIC-IV subset facilitates broader comparisons involving domain-specific and fine-tuned LLMs. All outcomes are aggregated into both Model-level and System-level metrics (right bracket).}
\label{fig:eval} \end{figure*}

Although evaluation of EHR-based annotation tools is challenging due to limited labeled truth, we performed a comprehensive evaluation of PrecLLM on both the local HNC dataset and the MIMIC-IV dataset (Table~\ref{tab:experimental_scope_final_plain}, Fig.~\ref{fig:eval}). The PrecLLM pipeline was designed to be flexible enough to accommodate varying formats and clinical questions tailored to each dataset. We conducted a comparative analysis by varying the LLM model, target phenotype, and preprocessing method; thus, both whole datasets and their subsets were used to evaluate different models. Because full-scope comparisons were time-consuming, part of the analysis was performed on a representative subset; broader comparisons with larger models were limited to the MIMIC-IV subset, and due to data privacy constraints, some fine-tuned models were evaluated only on MIMIC-IV. The two datasets shared some comparisons but required different evaluation considerations, and the experimental scope is summarized in Table~\ref{tab:experimental_scope_final_plain}. The PrecLLM procedure was motivated by and developed using the private HNC dataset, which contains rich real-time cancer notes (including early-stage cases), while MIMIC-IV provides validation and enables broader model/phenotype comparisons.

\begin{table}[ht] 
\centering 
\resizebox{\columnwidth}{!}{ 
\begin{tabular}{llcccccc} \toprule

\multicolumn{2}{l}{Dataset Configuration} & \makecell{Small\\LLM} & \makecell{Domain-\\Specific\\LLMs} & \makecell{Large\\LLM} & \makecell{Multi-Disease\\Phenotyping} & \makecell{Multi-Pre-\\processing\\Methods} & \makecell{Fine-Tuned\\LLM} \\ \midrule

\multirow{2}{*}{\makecell{Private UNC\\HNC Dataset}} & Subset & Yes & Yes & No\textsuperscript{a} & No\textsuperscript{c} & Yes & No\textsuperscript{d} \\ \cmidrule(l{4pt}r{4pt}){2-8} 
& Full Dataset & Yes & No\textsuperscript{b} & No\textsuperscript{a} & No\textsuperscript{c} & No\textsuperscript{b} & No\textsuperscript{d} \\ \midrule\midrule

\multirow{2}{*}{\makecell{MIMIC-IV\\Dataset}} & Subset & Yes & Yes & Yes & Yes & Yes & Yes \\ \cmidrule(l{4pt}r{4pt}){2-8} 
& Full Dataset & Yes & No\textsuperscript{b} & No\textsuperscript{b} & No\textsuperscript{b} & No\textsuperscript{b} & Yes \\ \bottomrule 
\end{tabular} }

\caption{Overview of the experimental scope and model configurations. The ``No'' entries correspond to specific feasibility constraints: 
(a) computational infeasibility of running large models on private hardware; 
(b) prohibitive cost of repeating all subset analyses on the full datasets; 
(c) insufficient reliable ICD codes for multi-disease analysis in the private HNC dataset; and 
(d) institutional privacy policies forbidding the use of private data for model fine-tuning. 
\textbf{Model Definitions:} Small LLM refers to \texttt{Gemma-7B-it}; Domain-Specific LLMs refer to \texttt{LLaMA-2-7B-Chat-Med} and \texttt{Bio-Medical-LLaMA-3-8B}; Large LLM refers to \texttt{Meta-Llama-3-70B-Instruct}; and Fine-tuned LLM refers to fine-tuned 2B or 7B models. Multi-preprocessing methods denote the use of multiple approaches, including RAG in addition to regex.} 
\label{tab:experimental_scope_final_plain} 
\end{table}

We emphasized the importance and complexity of evaluation, particularly at the visit level, for clinical decision-making and understanding disease progression, as decisions and progression were dynamic, time-sensitive, and longitudinal. We designed time-aware evaluations by constraining notes to specific care periods. 
In the private HNC dataset, the availability of the exact date of diagnosis with the ICD code for metastasis enabled time-sensitive evaluation at the visit level. To achieve that, given the ICD code as a ``true'' temporal anchor, we constructed a time window around this anchor point (e.g., 20, 30, or 40 days), selecting only the clinical notes recorded within this period for analysis. In this way, we assessed ``time-sensitive'' sensitivity at the visit level based on all these notes within that time window. 
However, the MIMIC-IV dataset, built mostly from patients in an intensive care unit (ICU), had intensive data in a short time. It had a different nature of the so-called visit level annotation. In fact, the hospital level of the MIMIC-IV dataset represented the visit level. In MIMIC-IV, the subject level evaluation also considered the situation where one patient might go to the ICU more than once.

The sensitivity of PrecLLM in both HNC and MIMIC-IV datasets was evaluated at both the subset and the whole-dataset levels, and at both the subject level and the visit level, using two different strategies (so-called Model vs System) to address the ``no notes found'' (NNF) cases in the evaluation metrics (Fig.~\ref{fig:eval}). 
The ground truth in evaluating PrecLLM sensitivity was defined by the corresponding ICD codes matched to the time period when the clinical notes were generated. In this evaluation, ICD codes served as a practical reference standard rather than definitive clinical truth.
Evaluation was organized in the following structure. Depending on how NNF was handled, sensitivity, specificity, and F1 could be reported at the model level or the system level (detailed in Method Section~\ref{sec:metrics}). 
The \textbf{model-level} metrics were calculated after excluding subjects for whom no relevant clinical notes were found. 
The \textbf{system-level} did not exclude any subjects; instead, cases where ``no notes were found'' for a real positive subject were considered a classification failure (i.e., a false negative).

It is worth noticing that sensitivity is typically more important than specificity in annotations of disease phenotypes, therefore the data subset was organized to enrich the target phenotypes (see subset definition in Data Description, Methods section). As a consequence of the enrichment, these subset datasets were more appropriate for evaluation of sensitivity but not for specificity (Fig.~\ref{fig:eval}). We only provided specificity in whole population analysis.
Furthermore, specificity at the HNC visit level (i.e., ``time-sensitive'') could not be a well-defined value. The basic unit of this analysis was the ``positive diagnosis event'' (i.e., existence of the ICD code with a timestamp), and all our calculations were centered around these. However, in the ``negative'' cohort, there was no such equivalent ``negative diagnosis event'' timestamp to serve as an anchor for creating our evaluation window, because we could not simply select a random time point from a negative patient as it did not constitute a comparable ``event''. Therefore, this ``time-sensitive'' methodology was inherently designed only to evaluate positive cases (sensitivity) and could not be applied in the same way to negative cases to calculate specificity.

\subsection{Preprocessing Enhances Classification Sensitivity of Metastasis in Private HNC Dataset}

\subsubsection{Overall Performance of Metastasis Classification in the Whole Private HNC Dataset}

We applied PrecLLM with \texttt{Gemma-7B-it} to the full private EHR dataset of 7,284 patients diagnosed with head and neck cancer (HNC) between 2014 and 2022 (See design in Table~\ref{tab:experimental_scope_final_plain}, Fig.~\ref{fig:eval}). In this full cohort ($n=7{,}284$), the grand truth is that 2,637 patients were identified as having metastasis based on their ICD codes, in the structured data. Among the 6,975 patients who have available clinical notes (1,737,908 total notes), 2,637 have ICD codes corresponding to metastasis in the structured data, considered true metastasis-positive in evaluation. 
 If we use regex in pre-filter, the metastasis-related keywords in notes can be identified from 4,844 patients, yielding 92,920 unique extracted notes for classification. 
 
\begin{figure*}[!t]
\centering

\begin{subfigure}[t]{0.49\textwidth}
  \centering
  \includegraphics[height=0.25\textheight]{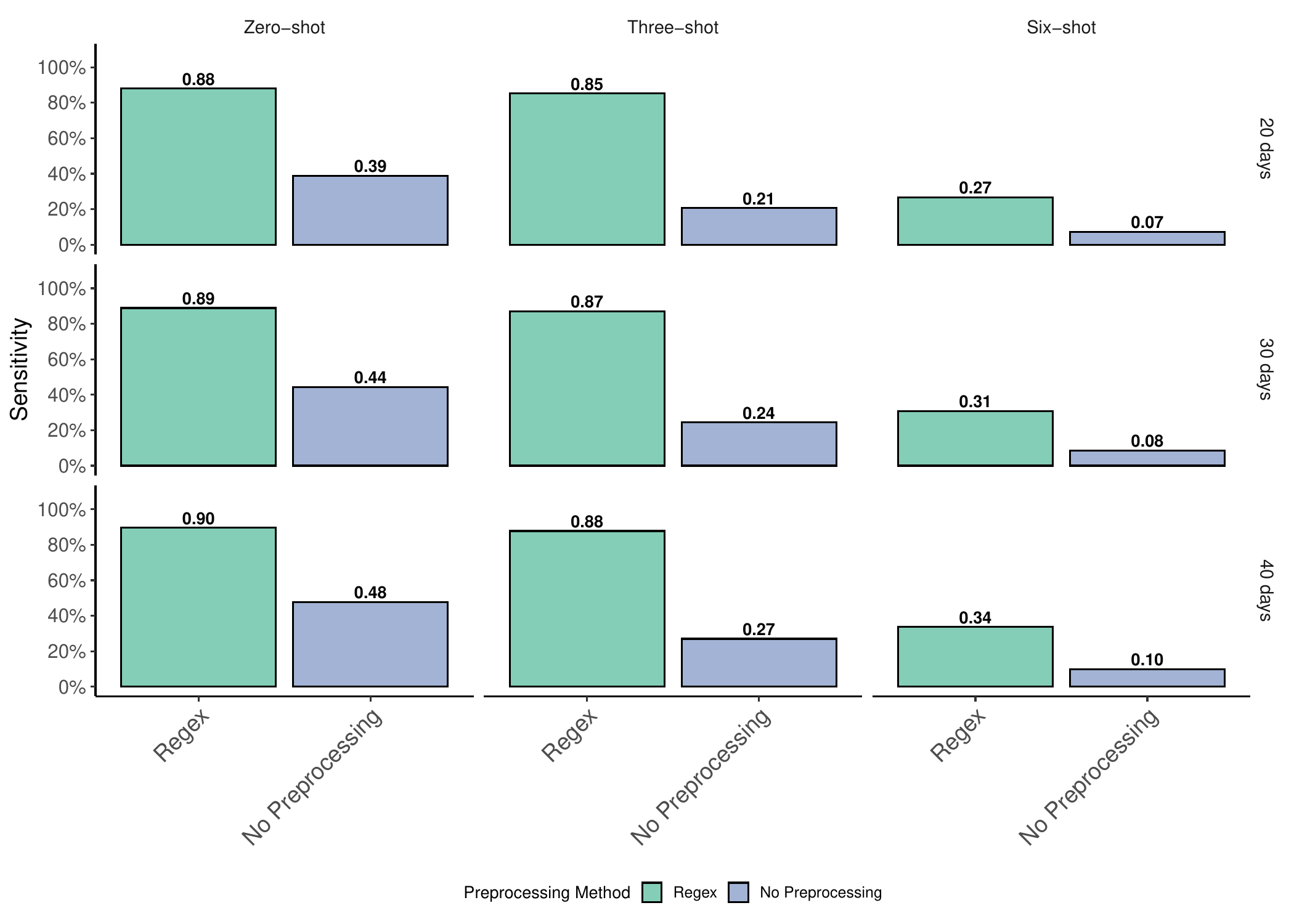}
  \caption{Whole private HNC: model sensitivity (visit level)}
  \label{fig:hnc_model_sens}
\end{subfigure}
\hfill
\begin{subfigure}[t]{0.49\textwidth}
  \centering
  \includegraphics[height=0.25\textheight]{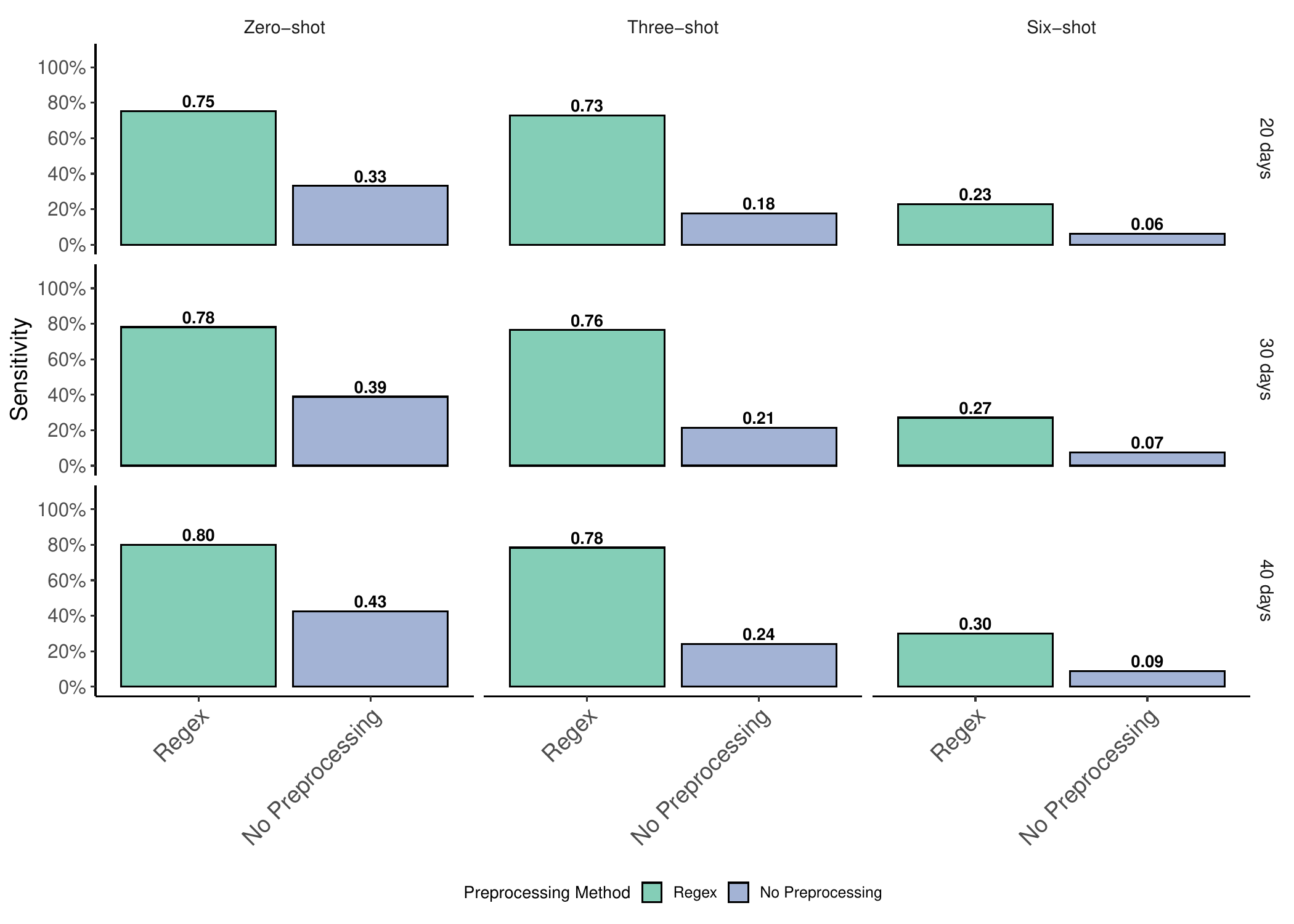}
  \caption{Whole private HNC: system sensitivity (visit level)}
  \label{fig:hnc_system_sens}
\end{subfigure}

\par\medskip

\begin{subfigure}[t]{0.49\textwidth}
  \centering
  \includegraphics[height=0.25\textheight]{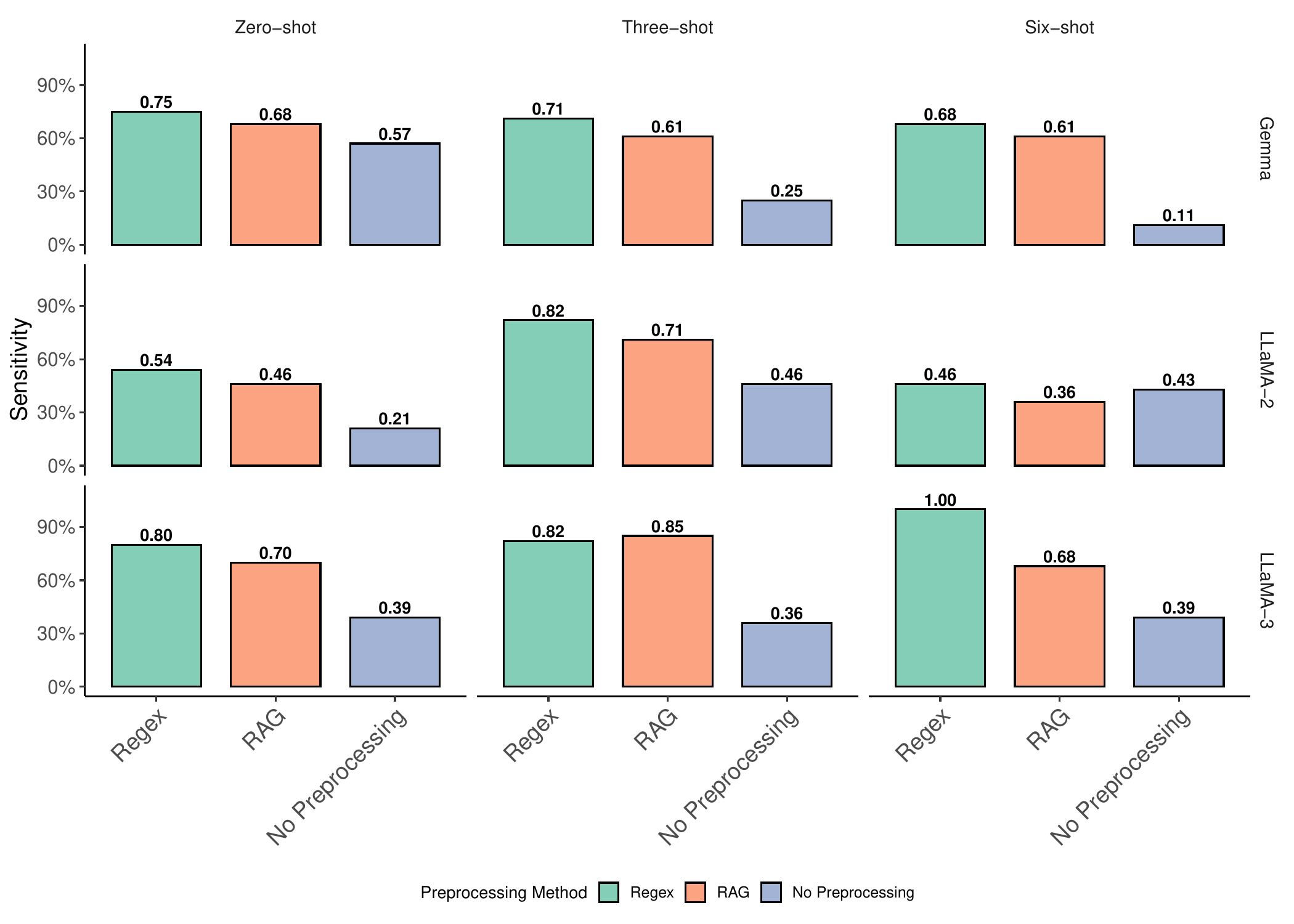}
  \caption{Private HNC subset: model sensitivity (subject level)}
  \label{fig:metastasis_subject_HNC}
\end{subfigure}
\hfill
\begin{subfigure}[t]{0.49\textwidth}
  \centering
  \includegraphics[height=0.25\textheight]{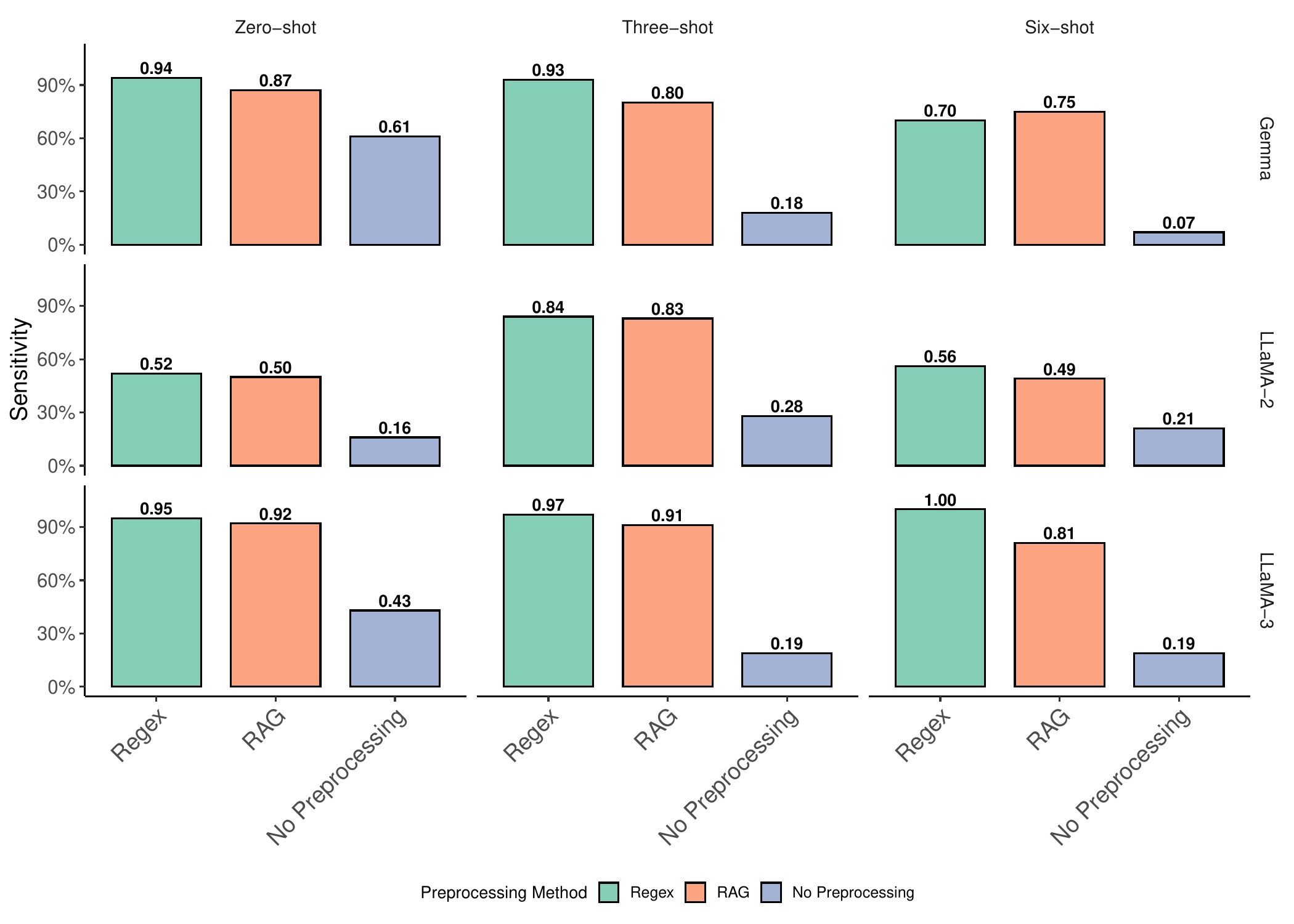}
  \caption{Private HNC subset: model sensitivity (visit level; 20-day window)}
  \label{fig:metastasis_20days}
\end{subfigure}

\par\medskip

\begin{subfigure}[t]{0.49\textwidth}
  \centering
  \includegraphics[height=0.25\textheight]{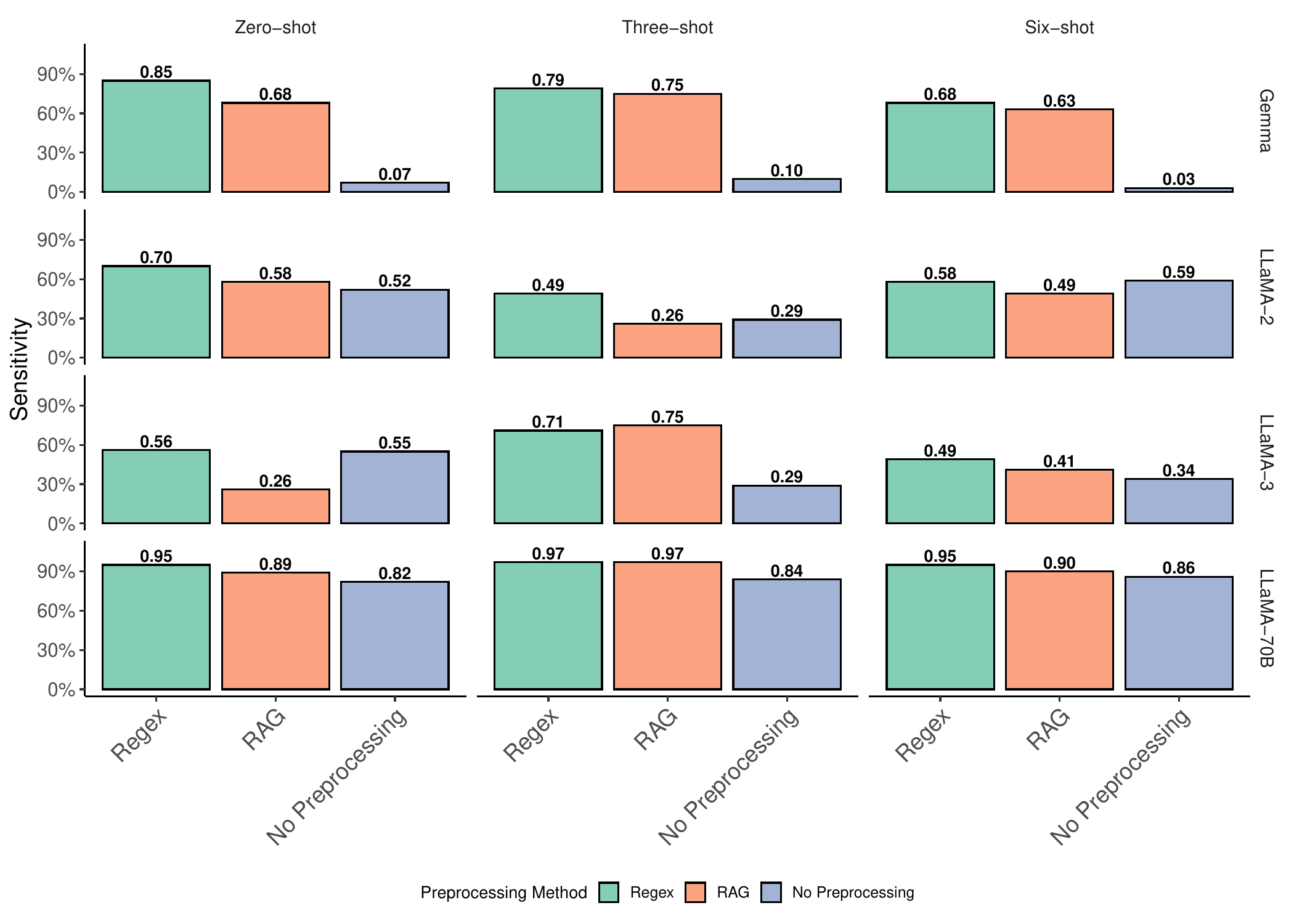}
  \caption{MIMIC-IV subset: model sensitivity (subject level)}
  \label{fig:metastasis_subject}
\end{subfigure}
\hfill
\begin{subfigure}[t]{0.49\textwidth}
  \centering
  \includegraphics[height=0.25\textheight]{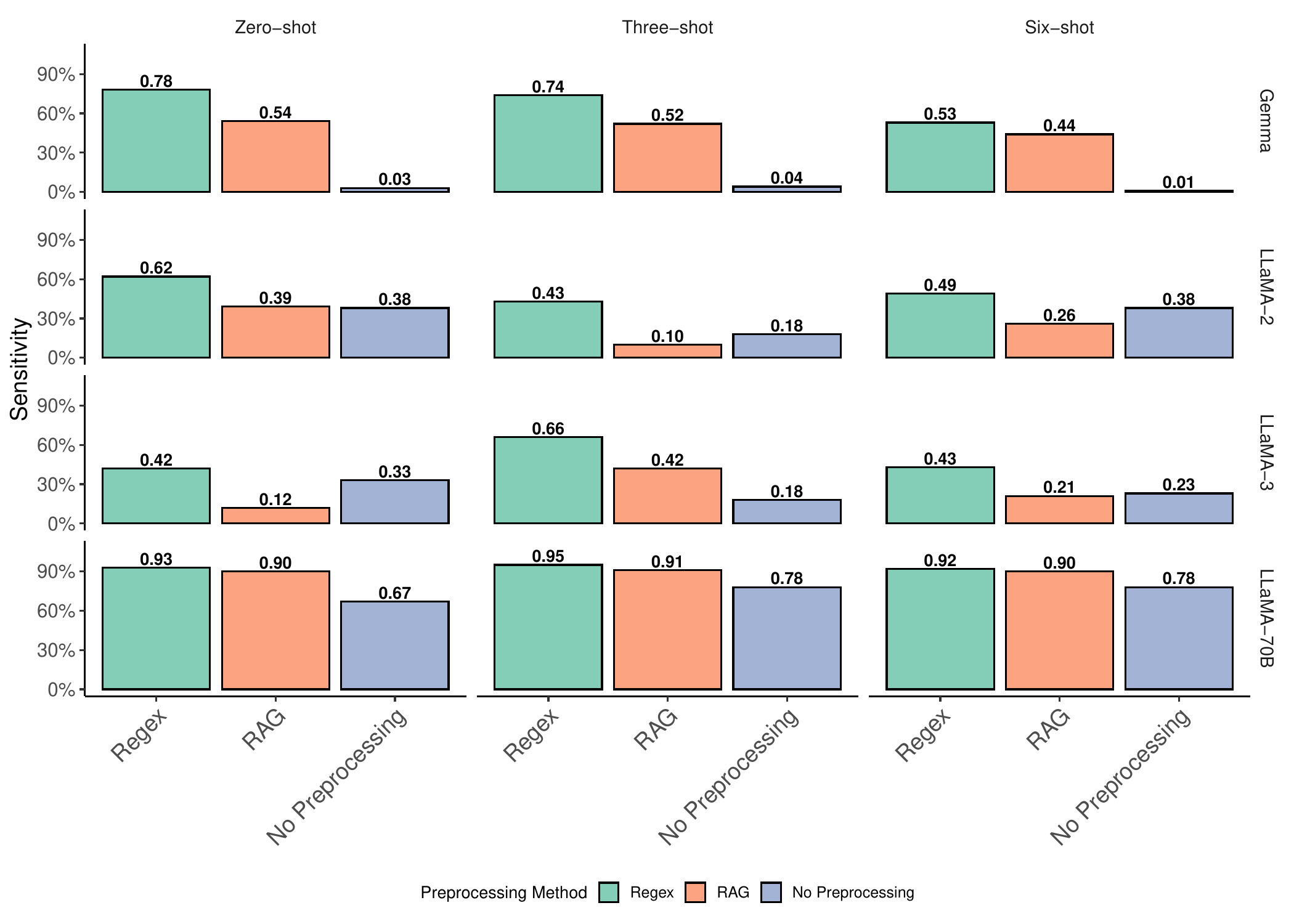}
  \caption{MIMIC-IV subset: model sensitivity (hospital admission level)}
  \label{fig:metastasis_hadm}
\end{subfigure}

\caption{Combined summary of metastasis-sensitivity analyses across three cohorts: the whole private HNC cohort, the private HNC subset, and the MIMIC-IV subset. All panels compare preprocessing settings (regex, RAG when applicable, and no preprocessing) under zero-, three-, and six-shot prompting.
(a) Whole private HNC (visit level, \texttt{Gemma-7B-it}): model sensitivity across 20/30/40-day windows centered on diagnosis dates.
(b) Same setting as (a): system sensitivity, counting missing-note cases as failures.
(c) Private HNC subset (subject level): model sensitivity for \texttt{Gemma-7B-it}, \texttt{LLaMA-2-7B-Chat-Med}, and \texttt{Bio-Medical-LLaMA-3-8B}.
(d) Private HNC subset (visit level, 20-day window): model sensitivity for the same three models.
(e) MIMIC-IV subset (subject level): model sensitivity for \texttt{Gemma-7B-it}, \texttt{LLaMA-2-7B-Chat-Med}, \texttt{Bio-Medical-LLaMA-3-8B}, and \texttt{Meta-Llama-3-70B-Instruct}.
(f) MIMIC-IV subset (hospital admission level): model sensitivity for the same four models.}

\label{fig:six_panel_sensitivity}
\end{figure*}

 Given the longitudinal nature of the clinical records, where note density and timing varied across patients, we evaluated phenotype annotation at two granularities: the aggregate subject level and the time-sensitive visit level. Specifically, we applied PrecLLM to notes collected within time windows around ICD diagnosis dates and then computed evaluation metrics (Methods Section 4.6). We carried out a comprehensive analysis of specificity alongside sensitivity at both the model and system levels. The results demonstrated the efficacy of preprocessing for both model sensitivity and system sensitivity at the subject level (Table~\ref{tab:hnc_summary_performance}), and for model sensitivity (Fig.~\ref{fig:hnc_model_sens}) and system sensitivity (Fig.~\ref{fig:hnc_system_sens}) at the visit level.

The detailed results demonstrated that preprocessing significantly enhanced overall classification performance across all learning paradigms at both the subject and visit levels. At the subject level (Table~\ref{tab:hnc_summary_performance}), this enhancement was particularly evident in the F1 score, reflecting a better balance of precision and recall. Preprocessing substantially boosted both Model F1 and model sensitivity. For instance, in the three-shot setting, the Model-level F1 increased from 0.5142 to 0.6912, accompanied by a sensitivity jump from 0.3858 to 0.7462. Similar significant gains in F1 scores were observed in the zero-shot (from 0.6112 to 0.6770) and six-shot (from 0.2745 to 0.5392) settings, with corresponding improvements in F1 and sensitivity in System metrics. These findings confirmed that preprocessing was critical for enhancing the overall reliability of the system when processing a subject's entire record. Consistent with this, at the visit level (Fig.~\ref{fig:hnc_model_sens} , \ref{fig:hnc_system_sens} ), regex-based preprocessing also improved sensitivity across prompting paradigms.
Notably, zero-shot prompting with preprocessing often outperformed few-shot settings. Increasing the number of shots in the small LLM \texttt{Gemma-7B-it} actually reduced sensitivity, suggesting the developed preprocessing to filter input in precLLM is more effective than prompt expansion. 
In addition, precLLM has shown robust performance of annotation extraction with different time windows of a visit. It was observed that the clinical notes within a time window of 20-days around a visit are sufficient to ensure sensitivity, compared to the longer windows of 30 or 40 days (Fig.~\ref{fig:hnc_model_sens}, \ref{fig:hnc_system_sens}).

\begin{table}[ht] \centering
\begin{tabular}{ccccccc} \toprule & \multicolumn{3}{c}{\textbf{Model Performance}} & \multicolumn{3}{c}{\textbf{System Performance}} \\ \cmidrule(lr){2-4} \cmidrule(lr){5-7} Experiment & Model Sens. & Model Spec. & Model F1 & System Sens. & System Spec. & System F1 \\ \midrule preprocessed, zero-shot & 0.7101 & 0.6065 & 0.6770 & 0.6568 & 0.7982 & 0.6517 \\ non-preprocessed, zero-shot & 0.5105 & 0.8378 & 0.6112 & 0.4721 & 0.9168 & 0.5829 \\ preprocessed, three-shot & 0.7462 & 0.5811 & 0.6912 & 0.6902 & 0.7852 & 0.6662 \\ non-preprocessed, three-shot & 0.3858 & 0.8835 & 0.5142 & 0.3568 & 0.9403 & 0.4878 \\ preprocessed, six-shot & 0.4190 & 0.8627 & 0.5392 & 0.3876 & 0.9296 & 0.5124 \\ non-preprocessed, six-shot & 0.1644 & 0.9659 & 0.2745 & 0.1521 & 0.9825 & 0.2571 \\ \bottomrule \end{tabular}

\caption{Performance Metrics for Metastasis Detection on the whole HNC Dataset (Subject Level) Based on \texttt{Gemma-7B-it}.} \label{tab:hnc_summary_performance} \end{table}

\subsubsection{Recall of Metastasis in the Private HNC EHR Data Subset}

Because LLM inference was computationally expensive, 
we applied PrecLLM to a randomly sampled 1\% subset of the real datasets to compare the performance of a larger number of small LLMs  combined  with two preprocessing strategies 
(regex and RAG) in the first and second steps of precLLM (See study design in Table~\ref{tab:experimental_scope_final_plain}, Fig.~\ref{fig:eval}), including \texttt{Gemma-7B-it}, \texttt{LLaMA-2-7B-Chat-Med}, and \texttt{Bio-Medical-LLaMA-3-8B}.
In the private HNC EHR dataset, we compare sensitivity of precLLM to annotate metastasis. Sensitivity here is also recall since the outcome is binary as Yes/No metastasis. There are 69 patients in this subset with their 23,875 clinical notes, as inpput to precLLM.  Among these 69 patients, 28 (40.6\%) had an ICD code for metastasis, and the prevalence was consistent with the whole dataset. In this subset, from the 1st step of precLLM, the regex procedure reduced the number of unique notes to 1,490 across 50 patients who have metastasis-related keywords, while RAG2 reduced to 9,836 unique notes for the 69 patients. 
For the evaluation result at the model-level, both regex and RAG substantially improved sensitivity over the non-preprocessed baseline for subjects (Fig.~\ref{fig:metastasis_subject_HNC}) and  for visits with notes from 20 days window (Fig.\ref{fig:metastasis_20days}), consistent with the evaluation for visits when 30 and 40 days are considered as the time window of clinical notes (SFigs.~\ref{fig:comparison_metastasis_HNC_3040}a and b in Supplementary Note~\ref{Sec:AddFig}).
For the evaluation results at the system-level (SFig.\ref{fig:comparison_sys_metastasis_HNC}). 
This pattern remained consistent across additional for subjects (SFig.\ref{fig:sys_metastasis_subject},) 
and time-window of 20, 30 and 40 days for visits (SFig.\ref{fig:sys_metastasis_20days}--\ref{fig:sys_metastasis_40days}). 
Similarly as in the whole HNC dataset, zero-shot prompting with preprocessing often matched or outperformed few-shot settings. Increasing the number of shots in these smaller models did not consistently improve performance but could reduce it, confirming that preprocessing in precLLM is more effective than the prompt expansion.

\subsection{Preprocessing Boosts Overall Performance of Multiple-Phenotype Annotation in MIMIC-IV }

\subsubsection{Overall Performance of Metastasis Classification in the Whole MIMIC-IV}

A comprehensive PrecLLM analysis of the full MIMIC-IV dataset including 145,914 patients (388,336 hospital admissions with 331,793 discharge notes) supported the main conclusions of this work (See design in Table~\ref{tab:experimental_scope_final_plain}, Fig.~\ref{fig:eval}). Of these, ICD codes indicated that 8,435 patients (5.78\%) had metastasis, and a regex search identified metastasis-related keywords in the notes of 13,923 patients (9.54\%) across 28,390 admissions (7.31\%), yielding 28,390 unique extracted notes (8.56\%). We compared preprocessing versus no preprocessing using \texttt{Gemma-7B-it} and fine-tuned 2B/7B models under zero-shot and few-shot prompts. As in the HNC dataset, the model and system evaluation metrics were demonstrated at both the subject level and the hospital admission level in the MIMIC-IV dataset. Given the significant computational resources such an exhaustive analysis requires, these particular tests were conducted using the above small models. The substantial impact of regex-based preprocessing is summarized in Tables \ref{tab:mimic_summary_performance_subject} and \ref{tab:mimic_summary_performance}.

As summarized in Tables \ref{tab:mimic_summary_performance_subject} and \ref{tab:mimic_summary_performance}, the results highlighted the substantial impact of regex-based preprocessing at both the subject-level and visit-level, i.e., hospital admission level in MIMIC. At the subject level, the upper block of Table~\ref{tab:mimic_summary_performance_subject} showed the same pattern observed in the first dataset, whole HNC data (Table~\ref{tab:hnc_summary_performance}). We observed that, preprocessing increased sensitivity and F1 (typically by 40\%--80\% in sensitivity across shot settings), with a moderate specificity trade-off. Fine-tuned models did not close this gap. Even the best fine-tuned result (23\% sensitivity, 34\% F1) remained well below zero-shot with preprocessing (87\% sensitivity, 84\% F1), although fine-tuning slightly improved the non-preprocessed baseline.  The hospital-admission analysis has shown the similar pattern and suggests the same conclusion (Table~\ref{tab:mimic_summary_performance}), indicating robustness across granularities and prompting settings.
Two additional trends were consistent. First, the performance in the small models often declined when more examples were included in the few-shot promp.  Second, zero-shot plus preprocessing outperformed fine-tuning, when a priori labeled data were limited which is typically the case in clinical unstructured data.

\begin{table}[ht] \centering

\resizebox{\columnwidth}{!}{

\begin{tabular}{ccccccc} \toprule & \multicolumn{3}{c}{\textbf{Model Performance}} & \multicolumn{3}{c}{\textbf{System Performance}} \\ \cmidrule(lr){2-4} \cmidrule(lr){5-7} Experiment & Model Sens. & Model Spec. & Model F1 & System Sens. & System Spec. & System F1 \\ \midrule

preprocessed, zero-shot & 0.8726 & 0.7026 & 0.8248 & 0.7852 & 0.9892 & 0.7836 \\ non-preprocessed,
zero-shot & 0.0568 & 0.9925 & 0.1069 & 0.0511 & 0.9997 & 0.0968 \\ preprocessed, three-shot & 0.6916
& 0.8150 & 0.7505 & 0.6223 & 0.9933 & 0.7078 \\ non-preprocessed, three-shot & 0.0166 & 0.9952 &
0.0325 & 0.0149 & 0.9998 & 0.0293 \\ preprocessed, six-shot & 0.4894 & 0.8828 & 0.6175 & 0.4404 &
0.9957 & 0.5770 \\ non-preprocessed, six-shot & 0.0048 & 0.9971 & 0.0096 & 0.0043 & 0.9999 & 0.0086
\\ \midrule preprocessed, Fine-tuned 2B & 0.2352 & 0.8172 & 0.3397 & 0.2116 & 0.9933 & 0.3144 \\
non-preprocessed, Fine-tuned 2B & 0.1929 & 0.7388 & 0.2743 & 0.1736 & 0.9905 & 0.2542 \\
preprocessed, Fine-tuned 7B & 0.1291 & 0.9018 & 0.2136 & 0.1162 & 0.9964 & 0.1956 \\
non-preprocessed, Fine-tuned 7B & 0.0008 & 0.9990 & 0.0016 & 0.0007 & 1.0000 & 0.0014 \\ \bottomrule

\end{tabular} }

\caption{Performance Metrics for Metastasis Detection on the whole MIMIC-IV Dataset (Subject Level) Based on \texttt{Gemma-7B-it}.} \label{tab:mimic_summary_performance_subject} \end{table}

\begin{table}[ht] \centering \resizebox{\columnwidth}{!}{ \begin{tabular}{ccccccc} \toprule & \multicolumn{3}{c}{\textbf{Model Performance}} & \multicolumn{3}{c}{\textbf{System Performance}} \\ \cmidrule(lr){2-4} \cmidrule(lr){5-7} Experiment & Model Sens. & Model Spec. & Model F1 & System Sens. & System Spec. & System F1 \\ \midrule

preprocessed, zero-shot & 0.8521 & 0.6381 & 0.7956 & 0.7237 & 0.9889 & 0.7348 \\ non-preprocessed,
zero-shot & 0.0270 & 0.9926 & 0.0522 & 0.0229 & 0.9998 & 0.0446 \\ preprocessed, three-shot & 0.5652
& 0.7955 & 0.6538 & 0.4801 & 0.9937 & 0.5929 \\ non-preprocessed, three-shot & 0.0074 & 0.9967 &
0.0147 & 0.0063 & 0.9999 & 0.0125 \\ preprocessed, six-shot & 0.3295 & 0.8854 & 0.4637 & 0.2799 &
0.9965 & 0.4122 \\ non-preprocessed, six-shot & 0.0019 & 0.9980 & 0.0038 & 0.0016 & 0.9999 & 0.0032
\\ \midrule preprocessed, Fine-tuned 2B & 0.2928 & 0.7675 & 0.3959 & 0.2487 & 0.9929 & 0.3535 \\
non-preprocessed, Fine-tuned 2B & 0.2811 & 0.7149 & 0.3724 & 0.2387 & 0.9913 & 0.3333 \\
preprocessed, Fine-tuned 7B & 0.1868 & 0.8716 & 0.2897 & 0.1586 & 0.9961 & 0.2546 \\
non-preprocessed, Fine-tuned 7B & 0.0013 & 0.9990 & 0.0025 & 0.0011 & 1.0000 & 0.0022 \\ \bottomrule
\end{tabular} }

\caption{Performance Metrics for Metastasis Detection on the whole MIMIC-IV Dataset (Hospital Admission Level) Based on \texttt{Gemma-7B-it}.} \label{tab:mimic_summary_performance} \end{table}

\subsubsection{Preprocessing Boosts Recall of Metastasis in the MIMIC-IV Subset}

Due to extensive computational requirements, we conducted sensitivity analyses on a random 1\% subset of MIMIC-IV subjects (Details in Method Section~\ref{Sec:dataset}). We selected a diverse model set (\texttt{Gemma-7B-it}, \texttt{LLaMA-2-7B-Chat-Med}, \texttt{Bio-Medical- LLaMA-3-8B}, and \texttt{Meta-Llama-3-70B-Instruct}) to test whether preprocessing consistently improved performance across architectures. Results in Figs.~\ref{fig:metastasis_subject} and \ref{fig:metastasis_hadm}, together with SFig.~\ref{fig:comparison_system_metastasis}, confirmed that preprocessing, na matter via regex or RAG, consistently yielded substantial gains over the non-preprocessed baseline where performance was minimal. In zero-shot at the subject level, \texttt{Gemma-7B-it} sensitivity increased from 7\% (raw) to 85\% (regex) and 68\% (RAG), with parallel system-sensitivity gains (6\% to 73\% and 59\%). Preprocessing also allowed smaller models to surpass larger raw-input baselines: \texttt{Gemma-7B-it} with regex (85\%) exceeded non-preprocessed \texttt{Meta-Llama-3-70B-Instruct} (82\%). While RAG provided a slight edge for the 70B model, regex remained highly effective across models. Consistent with HNC findings, adding more shots often degraded performance in smaller models, indicating that input-length control mattered more than prompt expansion. We further validated this on fine-tuned \texttt{Gemma-2B-it}/\texttt{Gemma-7B-it} models; as shown in SFig.~\ref{fig:finetune_sensitivity_comparison}, preprocessing remained more effective than fine-tuning under limited labeled data.

\begin{figure*}[ht]
    \centering
    \includegraphics[width=0.6\textwidth]{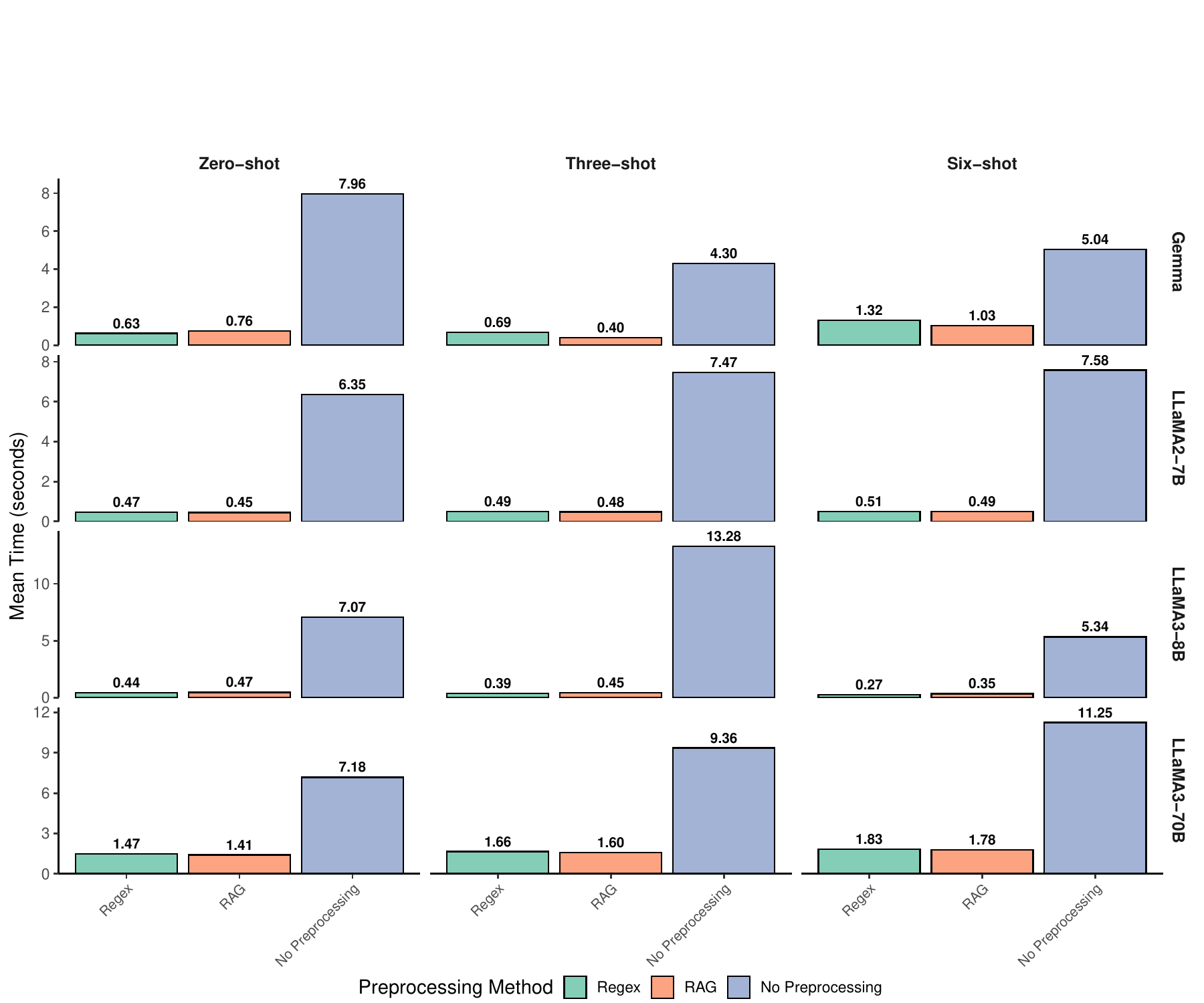}

    \caption{Mean LLM computation time per iteration for metastasis detection on the MIMIC-IV subset. The analysis was performed on a random sample of 100 subjects from the MIMIC-IV subset, evaluating computation time across different preprocessing methods (regex, RAG, and non-preprocessed), varying numbers of shots (zero-shot, three-shot, and six-shot), and different LLM models. Values are shown as mean seconds per iteration. The plot demonstrates that non-preprocessed methods were significantly more time-consuming than the regex and RAG methods, with preprocessing achieving over 20-fold speedup in many cases. \texttt{Gemma} = \texttt{Gemma-7B-it}, \texttt{LLaMA2-7B} = \texttt{LLaMA-2-7B-Chat-Med}, \texttt{LLaMA3-8B} = \texttt{Bio-Medical-LLaMA-3-8B}, \texttt{LLaMA3-70B} = \texttt{Meta-Llama-3-70B-Instruct}.}
    \label{fig:computation_time}
\end{figure*}

Beyond sensitivity, computational efficiency was a critical consideration for practical deployment. Comparison of the average computation time per iteration for each setting shows that the non-preprocessed approach was substantially slower,often by an order of magnitude (achieving a $>$20x speedup) than the regex and RAG methods (Fig.~\ref{fig:computation_time}). For example, in the three-shot setting, processing a single non-preprocessed note with \texttt{Bio-Medical-LLaMA-3-8B} took over 13 seconds on average, compared to less than half a second with regex. In summary, acorss all these diverse LLMs, the combination of higher sensitivity and markedly lower runtime highlights the  significant advantage to use PrecLLM for large-scale EHR annotation extraction in real clinical environment under constrained compute settings.

\subsubsection{Annotation Sensitivity of Insulin Use and Hypertension in MIMIC-IV Subset}

To demonstrate the generalizability and practical value of our framework, we presented additional results for Insulin use and Hypertension detection. The results for {model sensitivity} were shown in SFig.~\ref{fig:comparison_insulin} and SFig.~\ref{fig:comparison_hypertension}, while the corresponding {system sensitivity} results were in SFig.~\ref{fig:comparison_system_insulin} and SFig.~\ref{fig:comparison_system_hypertension}, respectively. Across both sets of metrics, these figures revealed the same fundamental patterns observed for metastasis: preprocessing was the most critical factor for achieving high model and system sensitivity, and the \texttt{Meta-Llama-3-70B-Instruct} model consistently outperformed the smaller models, especially when guided by a preprocessing step. This consistency across different clinical concepts validated the robustness of our proposed approach.

\FloatBarrier

\section{Discussion}

In this study, we introduce a novel compact LLM framework tailored for local deployment on EHR data, addressing significant challenges such as privacy concerns and computational limitations in healthcare settings. Our approach substantially enhances the performance of smaller LLMs by integrating preprocessing techniques, including but not limited to regex and RAG. This preprocessing not only reduces input complexity before it reaches the LLM but also ensures that the models focus on the most relevant information, thereby improving accuracy and efficiency. Moreover, preprocessing serves as a noise-reduction mechanism, helping to refine and standardize input data, ultimately leading to more reliable and interpretable model outputs. Furthermore, our pipeline offers inherent \textbf{traceability}: unlike ``black box'' approaches that process entire documents, PrecLLM isolates the specific text segments driving the decision, allowing clinicians to audit the AI's source of evidence. The efficacy of our framework is demonstrated through case studies on both private and public datasets, including the identification of ``metastasis'' in our private HNC dataset, as well as ``metastasis,'' ``insulin,'' and ``hypertension'' in the MIMIC-IV dataset. Unlike previous studies that have primarily focused on maximizing model performance without regard for computational cost, our work specifically targets the balance between performance and resource utilization, which is critical for practical healthcare applications. With the exception of tests involving the large-scale \texttt{Meta-Llama-3-70B-Instruct} model, all experiments presented in this paper can be conducted on \textit{a single NVIDIA 4090 GPU}. This minimal hardware requirement makes our approach feasible for many labs and may accelerate the adoption of LLM technology in biomedical research.

Our study provides practical insights into the local deployment of LLMs, particularly for smaller-scale models. All the compact LLMs we evaluate benefit substantially from preprocessing, highlighting its effectiveness in optimizing performance. The primary recommendation emerging from our work is to prioritize preprocessing as a more effective and resource-efficient strategy for enhancing model accuracy than relying solely on fine-tuning. This is supported by our experiments, which show that smaller models operating on properly preprocessed data produce better results than those fine-tuned with limited ground-truth data. Guidance on selecting a preprocessing technique can be based on the availability of domain knowledge. When some domain knowledge is available, simple methods like regex can be highly effective, providing a fast and efficient solution. Notably, this knowledge often requires limited domain expertise. Identifying relevant synonyms is usually sufficient, a task that can be easily handled using accessible GenAI tools. In cases where domain expertise is truly necessary, our study demonstrates that RAG can be a viable alternative. The utility of preprocessing is not restricted to smaller models; our findings confirm that this strategy also yields performance enhancements for larger models. A notable result from our study is that a compact model, augmented by an effective preprocessing procedure, can sometimes produce outcomes comparable to or better than those of a larger model working with non-preprocessed data. This observation supports preprocessing as a broadly applicable strategy across model scales.

While our findings are promising, our study has certain limitations. The framework's reliance on regex-based preprocessing may not fully capture the complexities of clinical language, potentially overlooking subtle yet clinically relevant information that falls outside predefined patterns. Although RAG-based methods can help address this limitation, their effectiveness can diminish without optimizing parameters, such as the number of retrieved sentences. As a potential future direction, investigating adaptive thresholding could enhance RAG’s effectiveness. Additionally, the effectiveness of our framework depends on the quality and diversity of the input data, which can vary across different healthcare settings, although we expect preprocessing to improve results in most scenarios compared to non-preprocessing. Future research could explore the integration of more advanced NLP techniques, such as context-aware embeddings or deep learning-based entity recognition, to further enhance the accuracy and adaptability of the preprocessing step. Moreover, leveraging existing biomedical databases or knowledge graphs could facilitate more advanced preprocessing, enriching the model's understanding of clinical concepts and relationships. Finally, assessing the framework across a broader range of diseases and clinical conditions would provide deeper insights into its generalizability and robustness.

\section{Methods}
\subsection{Introduction of Two EHR Datasets}
\label{Sec:dataset}

\paragraph{Private UNC HNC dataset} The private HNC dataset originated from the University of North Carolina (UNC) Health Care System. Data were retrieved from the Carolina Data Warehouse for Health (CDW-H), a centralized repository for UNC Health Care, in conjunction with the North Carolina Cancer Registry (NCCR), which was used to identify patients with confirmed cancer diagnoses. The specific cohort for this study included patients listed in the NCCR with a diagnosis of head and neck cancer between 2014 and 2022. Access to both CDW-H and NCCR data, including unstructured clinical notes containing PHI, was granted following approval of the Institutional Review Board (IRB). Data extraction and preparation were facilitated by the North Carolina Translational and Clinical Sciences Institute (NC TraCS), which acted as an impartial intermediary.

The private HNC dataset included a cohort of 7,284 patients. Clinical notes were available for 6,975 of these patients, totaling 1,737,908 notes. In this full cohort ($n=7{,}284$), 2,637 patients were identified as having
 metastasis based on their ICD codes in the structured data. Importantly among these 6,975 patients who have notes, 2,635 (37.8\%) had an ICD code for metastasis (definitions are provided in Supplementary Note~\ref{Sec:ICD_code} and Supplementary Tables~\ref{tab:icd_metastasis}--\ref{tab:icd_hypertension}). In addition, 4,844 patients (69.5\%) had notes containing metastasis-related keywords. This yielded 92,920 unique extracted notes, corresponding to 5.35\% of all 1,737,908 clinical notes from the 6,975 patients. We randomly sampled 1\% of 6,975 patients, consisting of 69 patients and their 23,875 clinical notes. Among 69 patients, 28 (40.6\%) had an ICD code for metastasis, and the prevalence was consistent with the whole dataset. Within this set of 69 patients, 50 patients with metastasis-related keywords were selected as a subset, yielding a total of 1,490 unique extracted notes (6.24\% of 23,875 notes of 69 patients). The cohort of these 50 patients was defined as the HNC subset and used for sensitivity evaluation.

\paragraph{MIMIC-IV dataset} The MIMIC-IV dataset was an extensive, anonymized EHR database encompassing data from patients admitted to Beth Israel Deaconess Medical Center between 2008 and 2019 \cite{Johnson2023MIMICIV}. This period covered more than a decade of ICU admissions, making it a widely used resource for medical research. With 26 tables of detailed patient data including demographics, lab measurements, procedures, medications, ICD codes, and clinical notes, MIMIC-IV offered comprehensive access to real-world clinical data. This database adhered strictly to the Health Insurance Portability and Accountability Act (HIPAA), ensuring patient privacy while supporting a wide range of medical studies. The breadth and depth of MIMIC-IV facilitated large-scale analyses relevant to clinical practice and health policy.

In the whole MIMIC-IV dataset, clinical discharge notes were available for 145,914 patients, corresponding to 388,336 hospital admissions and totaling 331,793 notes. Within this cohort, ICD codes indicated that 8,435 patients (5.78\%) had metastasis. A regex search for metastasis-related keywords found relevant text in the notes of 13,923 patients (9.54\%) across 28,390 admissions (7.31\%), resulting in a dataset of 28,390 unique extracted notes (8.56\%).

To evaluate sensitivity in subsets, we randomly sampled a 1\% set from the whole dataset, containing 1,459 patients, 3,785 hospital admissions, and 3,219 notes. This was for the target variables: metastasis, insulin use, and hypertension. The prevalence of these phenotypes in this new set was mostly consistent with the whole MIMIC-IV cohort, with 85 patients (5.83\%) having an ICD code for metastasis, 151 (10.35\%) for insulin use, and 817 (56.00\%) for hypertension, given that there were 8,435 patients (5.78\%) with ICD codes related to metastasis, 13,593 patients (9.32\%) with ICD codes related to insulin use and 81,685 patients (55.98\%) with ICD codes for hypertension in the whole dataset. The list of ICD codes used was provided in Supplementary Note~\ref{Sec:ICD_code}.

Specifically using a regex search for variable-related keywords, we identified 142 patients (9.73\%) across 265 admissions (7.00\%) for metastasis. We also found 286 patients (19.60\%) across 717 admissions (18.94\%) for insulin use, and 980 patients (67.17\%) across 2,214 admissions (58.49\%) for hypertension. These three subsets of 142, 286, or 980 patients were then used for sensitivity analysis for each of the three target variables.

\subsection{Experimental Design, Model Training, and Inference}

To evaluate the effectiveness of zero-shot learning, we crafted prompts to directly instruct the model to assess whether indications of the target phenotype were present based on the content of the provided text segment. This ensured that the model relied solely on its pre-existing knowledge and the context provided within the prompt without prior training. In our few-shot learning methodology, we adopted a demonstration-based in-context learning approach by selecting a balanced set of examples from the dataset. This collection typically consisted of $n$ positive samples, $n$ negative samples, and $n$ neutral samples. We termed this selection as ``$n$-shot'' learning. These samples were systematically integrated into the learning prompts to familiarize the LLMs with the specific nuances of the task. We listed the prompts for both zero-shot and few-shot learning in the Supplementary Note~\ref{Sec:prompt}.

To compare the effectiveness of our preprocessing-based approach with traditional optimization, we fine-tuned pre-trained LLMs, including \texttt{Gemma-2B} and \texttt{Gemma-7B}. Given the absence of definitive note-level ground truth in the MIMIC-IV dataset--primarily because it provided only subject level ICD codes without timestamps for individual notes--we manually curated a specific dataset for this fine-tuning process. This involved selecting 50 samples explicitly containing metastasis-related terms as positive examples and 50 samples clearly indicating the absence of metastatic conditions as negative examples. These samples were chosen randomly, with the possibility of multiple samples originating from the same patient. We then split these 100 curated samples into two sets: 80\% of the samples were used for training the model, while the remaining 20\% were set aside for validation. To enhance the model's adaptability without extensively altering its pre-trained weights, we incorporated the Low-Rank Adaptation (LoRA,~\cite{hu2021lora}) technique. The trainable parameters were around 3.04\% for \texttt{Gemma-2B} and 2.29\% for \texttt{Gemma-7B}.

In the final experimental inference procedure, the designed prompts (or fine-tuned models) were applied to the text segments generated by the PrecLLM framework. For the non-preprocessed baseline, the entire clinical note served as the input context, truncated only when exceeding the model's context window, to benchmark the efficacy of the filtering step. We then systematically evaluated classification performance across these different input settings (preprocessed vs. non-preprocessed) and various model architectures.

\subsection{Majority Vote to Determine the Status of the Target Variable} \label{sec:majority_vote} Following the previously described procedures, the LLM assigned a classification status to each clinical note, regardless of whether it was subjected to preprocessing. This status, determined through task-specific prompts, was categorized as ``Yes'' (indicating the presence of the target phenotype), ``No'' (indicating its absence), or ``Unknown''. The ``Unknown'' category was essential for appropriately classifying clinical notes that did not contain definitive information regarding the phenotype's status. For example, a note might indicate that a clinician had conducted examinations or ordered diagnostic tests for which the results were pending, thereby precluding an immediate confirmation or negation of the phenotype.

To ascertain the overall status of the target phenotype for a specific subject or hospital admission over a defined period, we applied a majority voting rule to all LLM-generated classifications pertaining to that subject or admission. Specifically, if the number of ``Yes'' classifications was greater than the number of ``No'' classifications, the subject or admission was determined to be ``Yes'' for the phenotype. Conversely, if ``No'' classifications were more numerous, the status was determined as ``No''. In cases where the counts of ``Yes'' and ``No'' classifications were equal, or if all classifications were ``Unknown'', the final status for the subject or admission was designated as ``Undecided''. It was noteworthy that when employing regex for preprocessing, a single clinical note might yield multiple instances of keyword-related text segments. Each such segment could be individually processed by the LLM, potentially resulting in multiple status classifications originating from one initial clinical note. For the purpose of determining the final status of the subject or hospital admission, each of these individual classifications was regarded as a distinct input to the majority voting process.

\subsection{Evaluation Metrics} \label{sec:metrics}

\subsubsection{Evaluation metrics at the subject level in both HNC and MIMIC-IV datasets, and at the hospital admission level in MIMIC-IV} \label{sec:evaluation_metrics_1}

First, we established the reference standard (serving as a proxy for ground truth) using ICD codes. Let $M_{ICD}$ be the group of units where the condition was present, and $N_{ICD}$ be the group where it was absent. Units could be subjects (in both HNC and MIMIC-IV) or hospital admissions (only in MIMIC-IV). At the subject level, $M_{ICD}$ included those who had at least one phenotype ICD code at any time. At the hospital admission level, $M_{ICD}$ included admissions that had at least one phenotype ICD code at any time. Here, ``NNF'' was a possible output by regex when no phenotype-related clinical notes were found in the first step of our pipeline before using the LLM (e.g., note 3 on the left of Step 1 in Figure~\ref{fig:detail}), denoted as $N_{NNF}$.

Second, we defined the positives and negatives from PrecLLM. For each unit that had the related clinical notes, one of the three outcomes ``Yes'', ``No'', or ``Undecided'' was assigned through majority vote in later steps based on the unit's clinical notes (see Section~\ref{sec:majority_vote}). We then had {$M_{LLM}$} as the set of units classified as ``Yes''. Since both ``No'' and ``Undecided'' suggested that the phenotype to be annotated was not identified, for simplicity, we grouped them together as the non-positive, ``No'' set {$N_{LLM}$}.

Third, in the performance metrics, there were two different ways to count the units assigned as NNF, depending on whether the evaluation was in the aspect of evaluating model performance (``model'') or evaluating the performance of the whole annotation system (``system''). In the model evaluation, Model Performance assessed the LLM's capability only on the units that had related clinical notes (therefore NNF was excluded in evaluation). This set of units, the evaluated cohort, included the following two cohorts as the ground truth positives and negatives. After excluding NNF, the ``evaluated positive'' cohort, {$M_{ICD, eval}$}, was the set of ground truth positive units for which an ICD code of the phenotype was found ($M_{ICD} - N_{NNF}$); similarly, the ``evaluated negative'' cohort, {$N_{ICD, eval}$}, was the set of ground truth negative units for which an ICD code of the phenotype was not found ($N_{ICD} - N_{NNF}$). Therefore, {model sensitivity} and {Model Specificity} were then calculated as $ \text{Sens}_{\text{Model}} = \frac{\#(M_{LLM} \cap M_{ICD, eval})}{\#(M_{ICD, eval})} $ and $ \text{Spec}_{\text{Model}} = \frac{\#(N_{LLM} \cap N_{ICD, eval})}{\#(N_{ICD, eval})} $. In the system evaluation, the System performance metrics measured the real-world success of the \textit{entire pipeline} of PrecLLM, from finding relevant notes to correctly classifying them. To achieve this, the total positive cohort ($M_{ICD}$) and negative cohort ($N_{ICD}$) were used for calculating the System Performance. These metrics used the complete, original cohorts ($M_{ICD}$ and $N_{ICD}$). Critically, a ``NNF'' ($N_{NNF}$) outcome was handled as part of the performance calculation. A ``NNF'' outcome for a unit in $M_{ICD}$ was a pipeline failure and was counted as a {System False Negative}. A ``NNF'' outcome for a unit in $N_{ICD}$ was a correct pipeline rejection and was counted as a {System True Negative}. {system sensitivity} and {System Specificity} were then calculated on the \textit{total} population, which could be expressed as $ \text{Sens}_{\text{System}} = \frac{\#(M_{LLM} \cap M_{ICD})}{\#(M_{ICD})} $ and $ \text{Spec}_{\text{System}} = \frac{\#((N_{LLM} \cup N_{NNF}) \cap N_{ICD})}{\#(N_{ICD})} $.

Finally, to provide a balanced metric, we calculated the F1-score, defined as the harmonic mean of sensitivity and precision. We defined precision as the proportion of correctly predicted positive units among all units classified as positive: $\text{Prec} = \frac{\#(M_{LLM} \cap M_{ICD})}{\#(M_{LLM})}$. Note that precision remains formulationally consistent across both Model and System evaluations. Consequently, the F1-score was calculated as
$F1 = 2 \cdot \frac{\text{Sens} \cdot \text{Prec}}{\text{Sens} + \text{Prec}}
$, where $\text{Sens}$ corresponds to either $\text{Sens}_{\text{Model}}$ or $\text{Sens}_{\text{System}}$.

We applied this evaluation framework according to the granularities available in each dataset. For the HNC dataset, the analysis was conducted at the subject level. In contrast, for the MIMIC-IV dataset, we evaluated at both the per-individual (subject) level and the per-hospital admission (visit) level. The subject-level analysis provided a single, cumulative classification based on the patient's entire medical history, while the admission-level analysis yielded distinct, time-sensitive classifications for each separate hospital encounter.

\subsubsection{Evaluation of ``Time-sensitive'' sensitivity at the visit level in the private HNC dataset}

For the private HNC dataset, we evaluated a ``time-sensitive'' sensitivity, which was made possible by the inclusion of precise timestamps for each diagnosis entry (i.e., the ICD code of the target variable). Each of these timestamped entries was treated as a distinct ground truth event for this evaluation.

From the output of PrecLLM, we first identified all patients who had at least one ICD code for the target variable (e.g., metastasis), previously defined as the positive set $M_{ICD}$. For each patient $i$ in this positive cohort (where $i = 1, \dots, \#M_{ICD}$), let $t_{ij}$ denote the time of the $j$-th diagnosis event of the target variable ICD for that patient. For each such event time $t_{ij}$, a corresponding visit level annotation (derived from a time window) was generated as part of the PrecLLM output. This visit level of annotation was based on the clinical notes found within a predefined $n$-day window centered on the target variable diagnosis date (e.g., $\frac{n}{2}$ days before and after $t_{ij}$). Therefore, each such ICD event had one annotation result ($A_{ij}$), one of the four ``Yes'', ``No'', ``Undecided'' and ``NNF''. In our analysis, three time windows ($n=20$, $30$, and $40$) were checked for comparison.

We defined $N_c = \sum_{i, j} I(A_{ij} = \text{``Yes''})$ as the count of correct outcomes, where ``Yes'' was classified by the LLM based on majority vote in PrecLLM for all notes in the time window; $N_w = \sum_{i, j} I(A_{ij} = \text{``No''} \text{ or } A_{ij} = \text{``Undecided''})$ as the count of outcomes with an incorrect outcome (``No'') or with an inconclusive outcome (``Undecided''). An additional count, $N_{nn} = \sum_{i, j} I(A_{ij} = \text{``NNF''})$, denotes the number of outcomes for whom no relevant notes were found within the specified time window. Using these counts, we defined two metrics. The {time-sensitive model sensitivity}, which evaluated performance only on cases where notes were available, was defined as $Sens_{Model} = \frac{N_{c}}{N_c + N_w}$. The time-sensitive system sensitivity, which provides an end-to-end assessment by penalizing missing notes, was defined as $Sens_{System} = \frac{N_{c}}{N_c + N_w + N_{nn}}$.

\bibliographystyle{unsrt} \bibliography{refs.bib}

\clearpage
\appendix
\section*{Supplementary Information}
\setcounter{section}{0}
\renewcommand{\thesection}{\arabic{section}}
\setcounter{figure}{0}
\renewcommand{\thefigure}{\arabic{figure}}
\renewcommand{\figurename}{Supplementary Fig.}
\setcounter{table}{0}
\renewcommand{\thetable}{\arabic{table}}
\renewcommand{\tablename}{Supplementary Table}

\section{Additional Figures}
\label{Sec:AddFig}
\begin{figure*}[ht]
    \centering
    \begin{subfigure}[b]{0.49\textwidth}
        \centering
        \includegraphics[width=\textwidth]{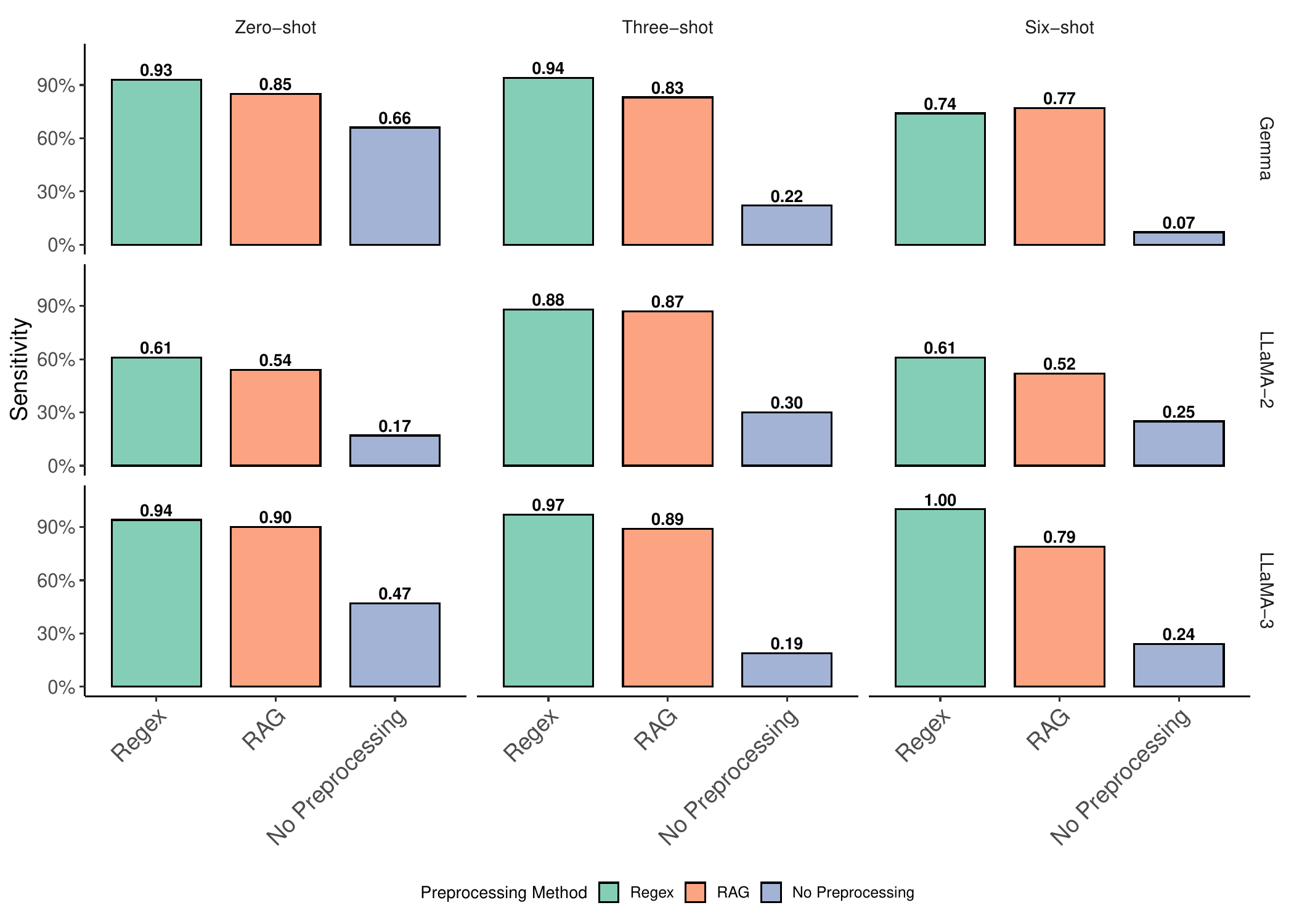}
        \caption{Model sensitivity for metastasis detection within a 30-day time window on the HNC subset.}
        \label{fig:metastasis_30days}
    \end{subfigure}
    \hfill
    \begin{subfigure}[b]{0.49\textwidth}
        \centering
        \includegraphics[width=\textwidth]{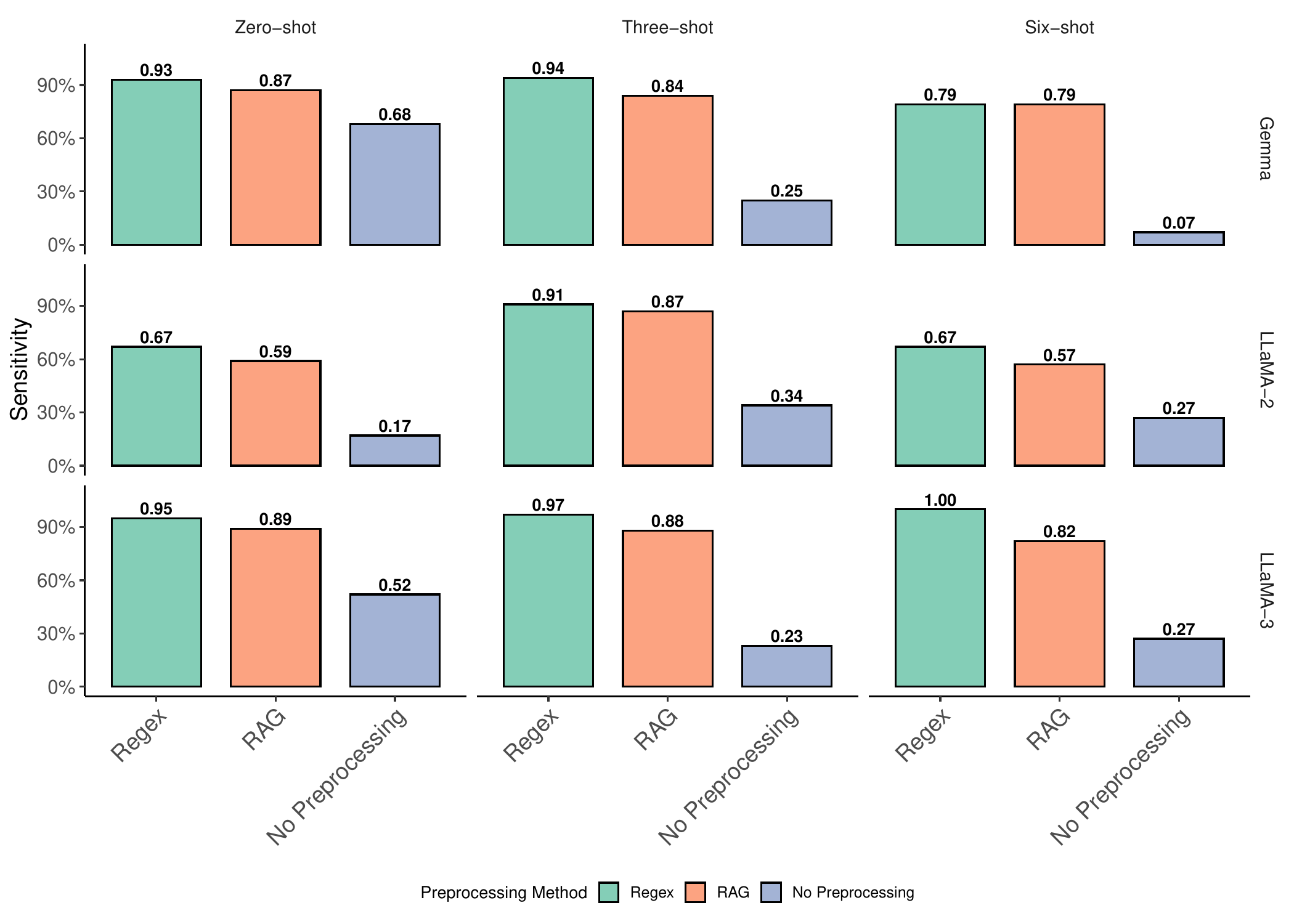}
        \caption{Model sensitivity for metastasis detection within a 40-day time window on the HNC subset.}
        \label{fig:metastasis_40days}
    \end{subfigure}

    \caption{Model sensitivity for metastasis detection on the private HNC dataset subset, evaluated under zero-shot, three-shot, and six-shot settings with regex, RAG, and non-preprocessed methods across 30-day and 40-day time windows.}
    \label{fig:comparison_metastasis_HNC_3040}
\end{figure*}

\begin{figure*}[ht]
    \centering
    \begin{subfigure}[b]{0.49\textwidth}
        \centering
        \includegraphics[width=\textwidth]{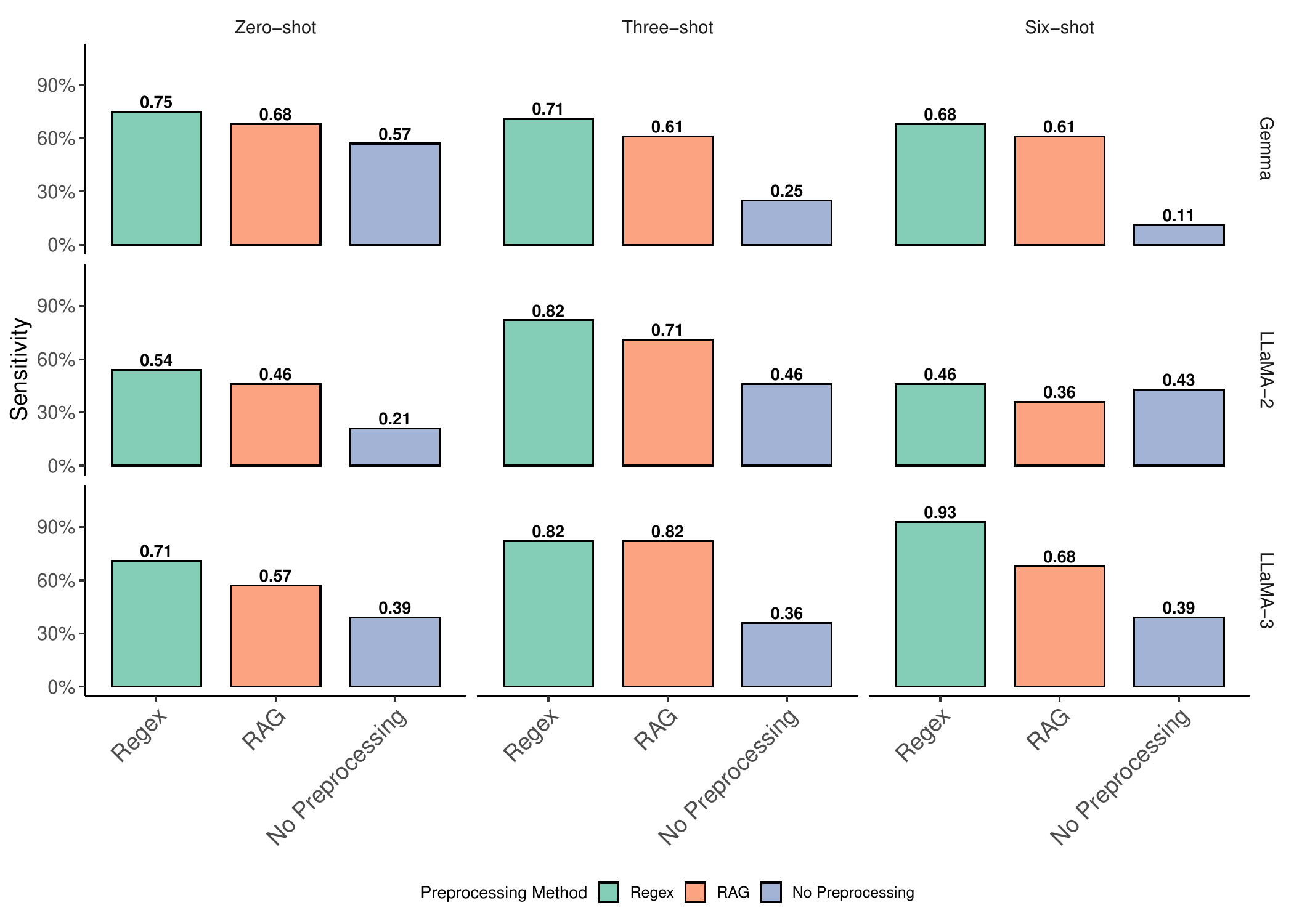}
        \caption{System sensitivity for metastasis detection at the subject level on the HNC subset.}
        \label{fig:sys_metastasis_subject}
    \end{subfigure}
    \hfill
    \begin{subfigure}[b]{0.49\textwidth}
        \centering
        \includegraphics[width=\textwidth]{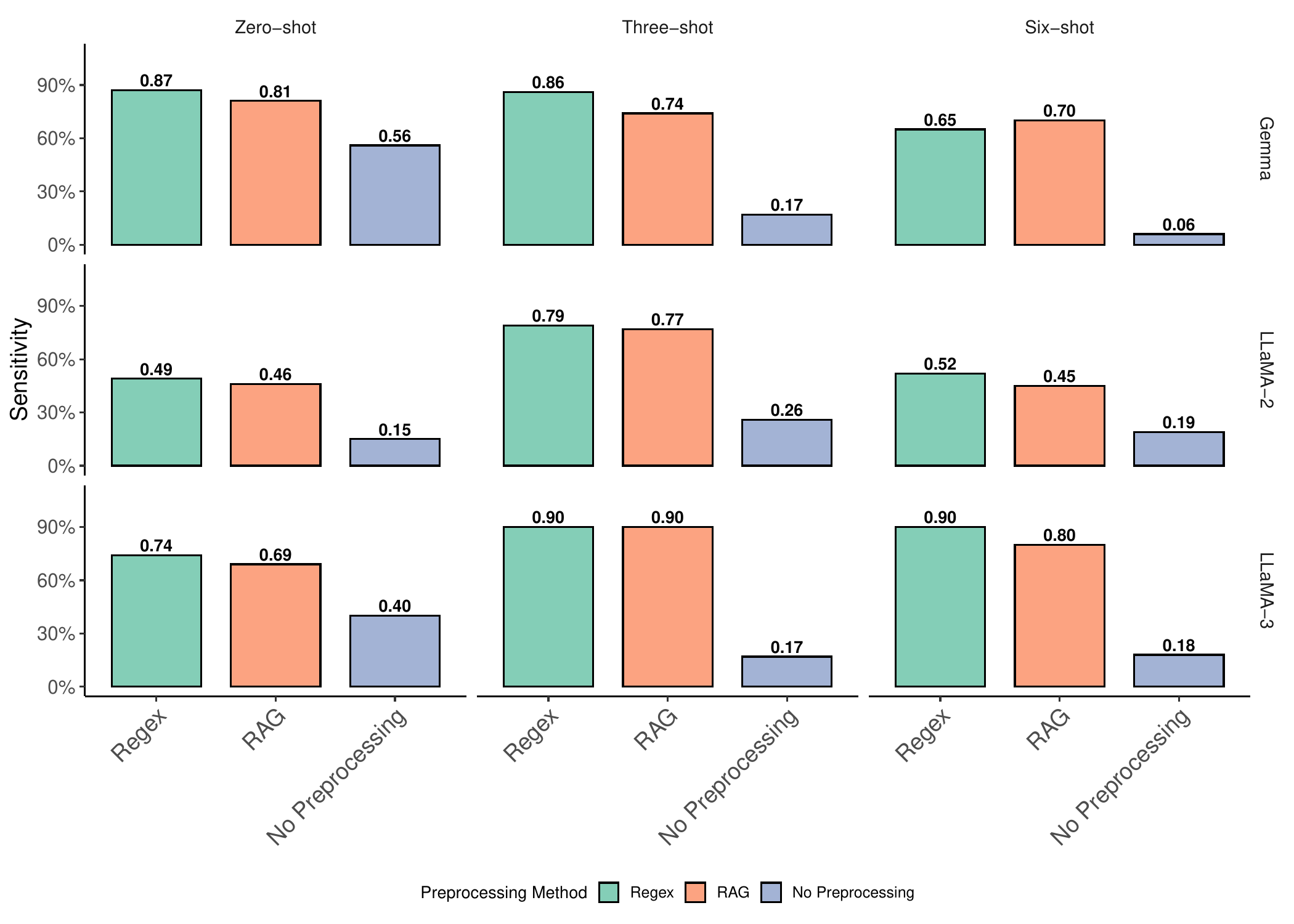}
        \caption{System sensitivity for metastasis detection within a 20-day time window on the HNC subset.}
        \label{fig:sys_metastasis_20days}
    \end{subfigure}

    \vspace{1em}

    \begin{subfigure}[b]{0.49\textwidth}
        \centering
        \includegraphics[width=\textwidth]{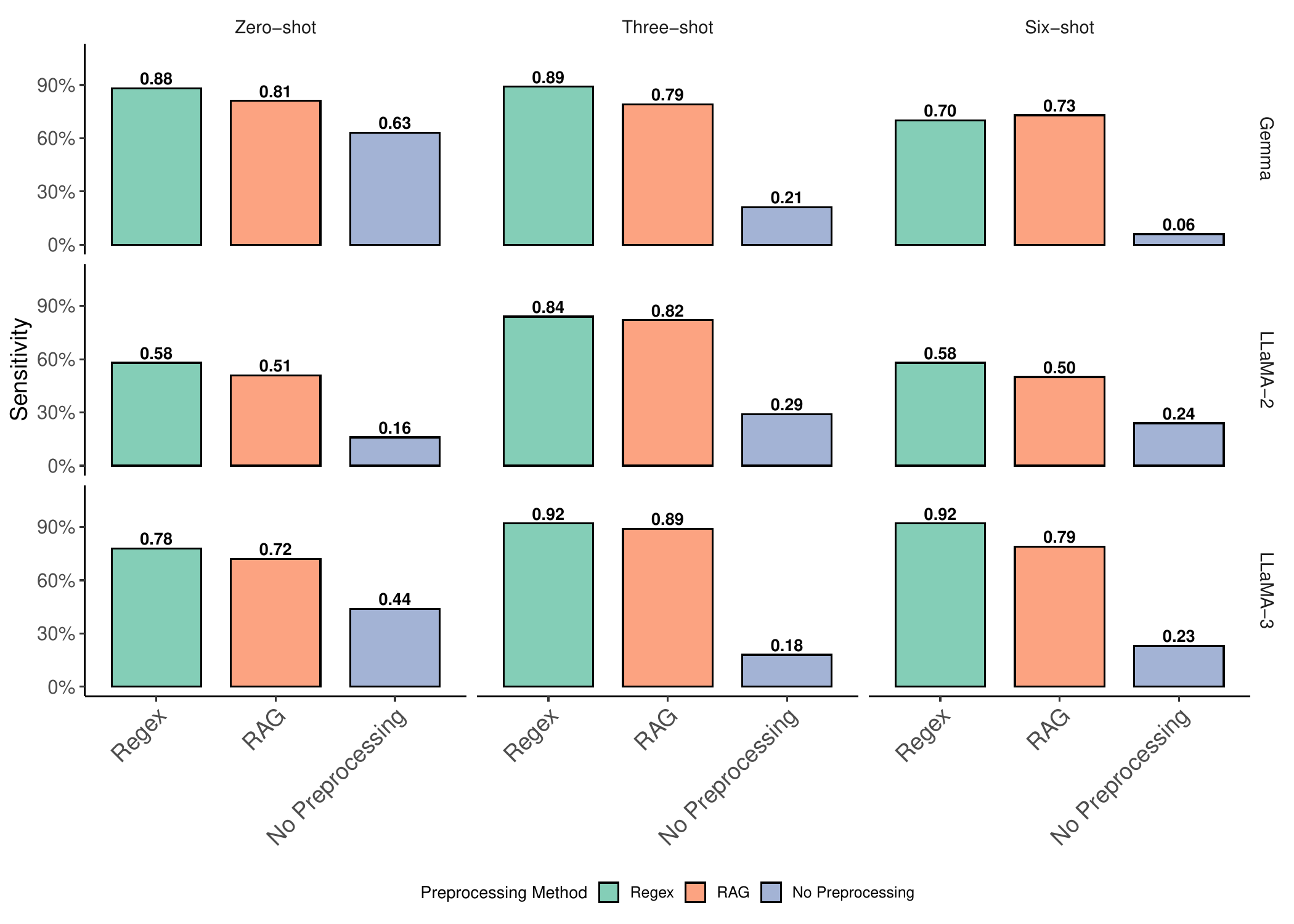}
        \caption{System sensitivity for metastasis detection within a 30-day time window on the HNC subset.}
        \label{fig:sys_metastasis_30days}
    \end{subfigure}
    \hfill
    \begin{subfigure}[b]{0.49\textwidth}
        \centering
        \includegraphics[width=\textwidth]{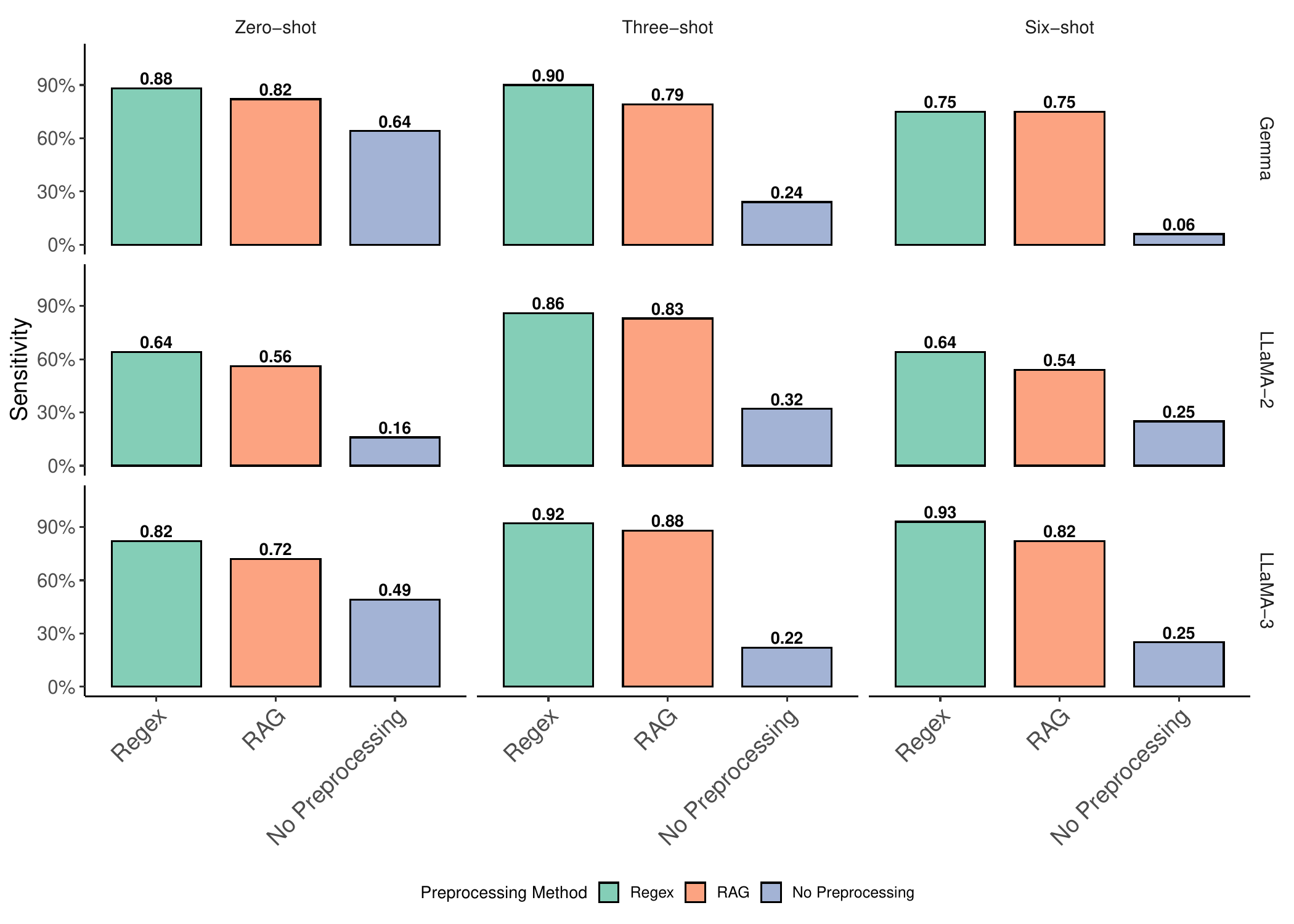}
        \caption{System sensitivity for metastasis detection within a 40-day time window on the HNC subset.}
        \label{fig:sys_metastasis_40days}
    \end{subfigure}

    \caption{System sensitivity for metastasis detection on the private HNC dataset subset at the subject level and visit level across 20, 30, and 40-day time windows, evaluated under zero-shot, three-shot, and six-shot settings with regex, RAG, and non-preprocessed methods.}
    \label{fig:comparison_sys_metastasis_HNC}
\end{figure*}

\begin{figure*}[ht]
    \centering

    \begin{subfigure}[b]{0.49\textwidth}
        \centering
        \includegraphics[width=\textwidth]{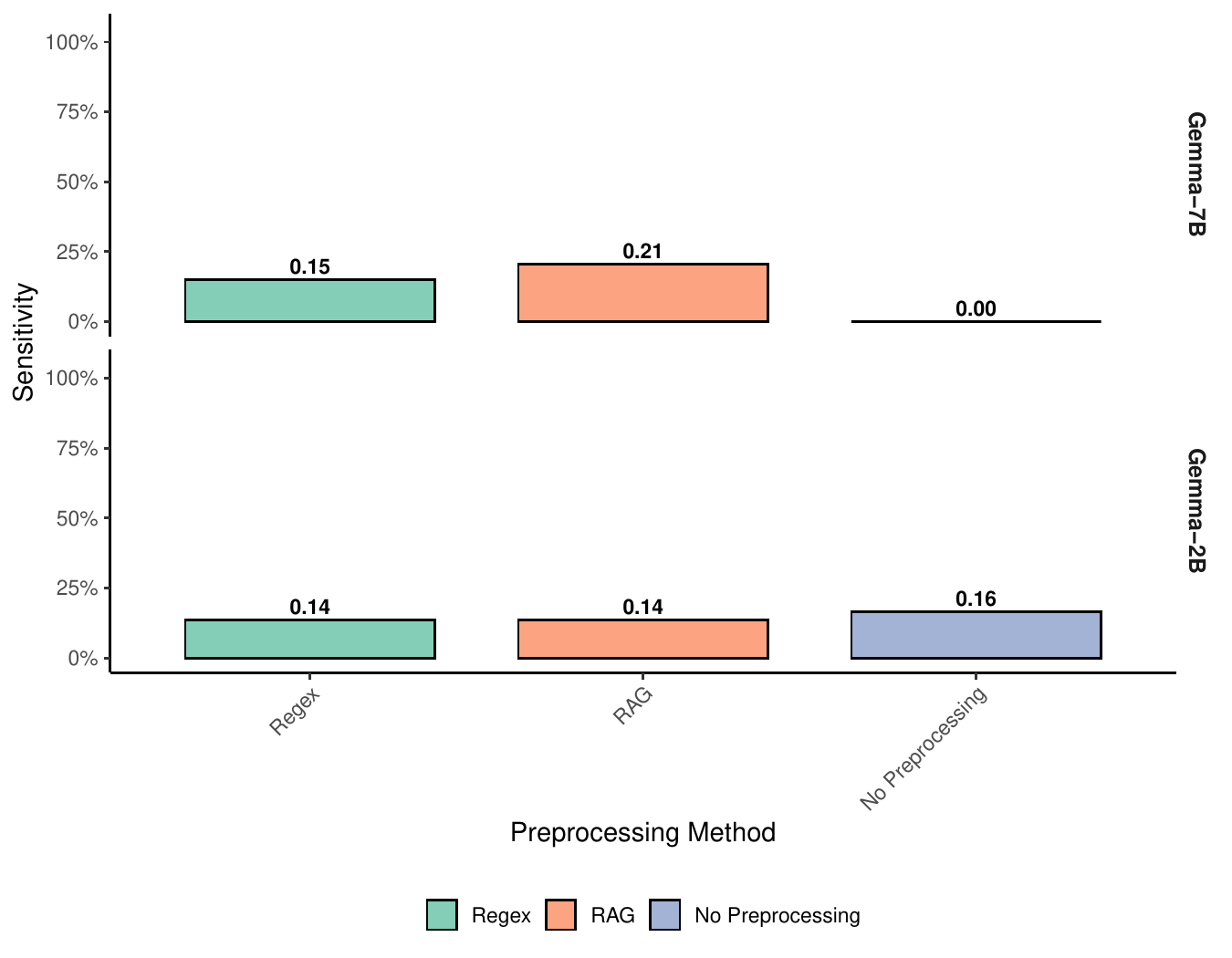}
        \caption{Model sensitivity for metastasis detection at the subject level using fine-tuned Gemma models.}
        \label{fig:subject_model_sens}
    \end{subfigure}
    \hfill
    \begin{subfigure}[b]{0.49\textwidth}
        \centering
        \includegraphics[width=\textwidth]{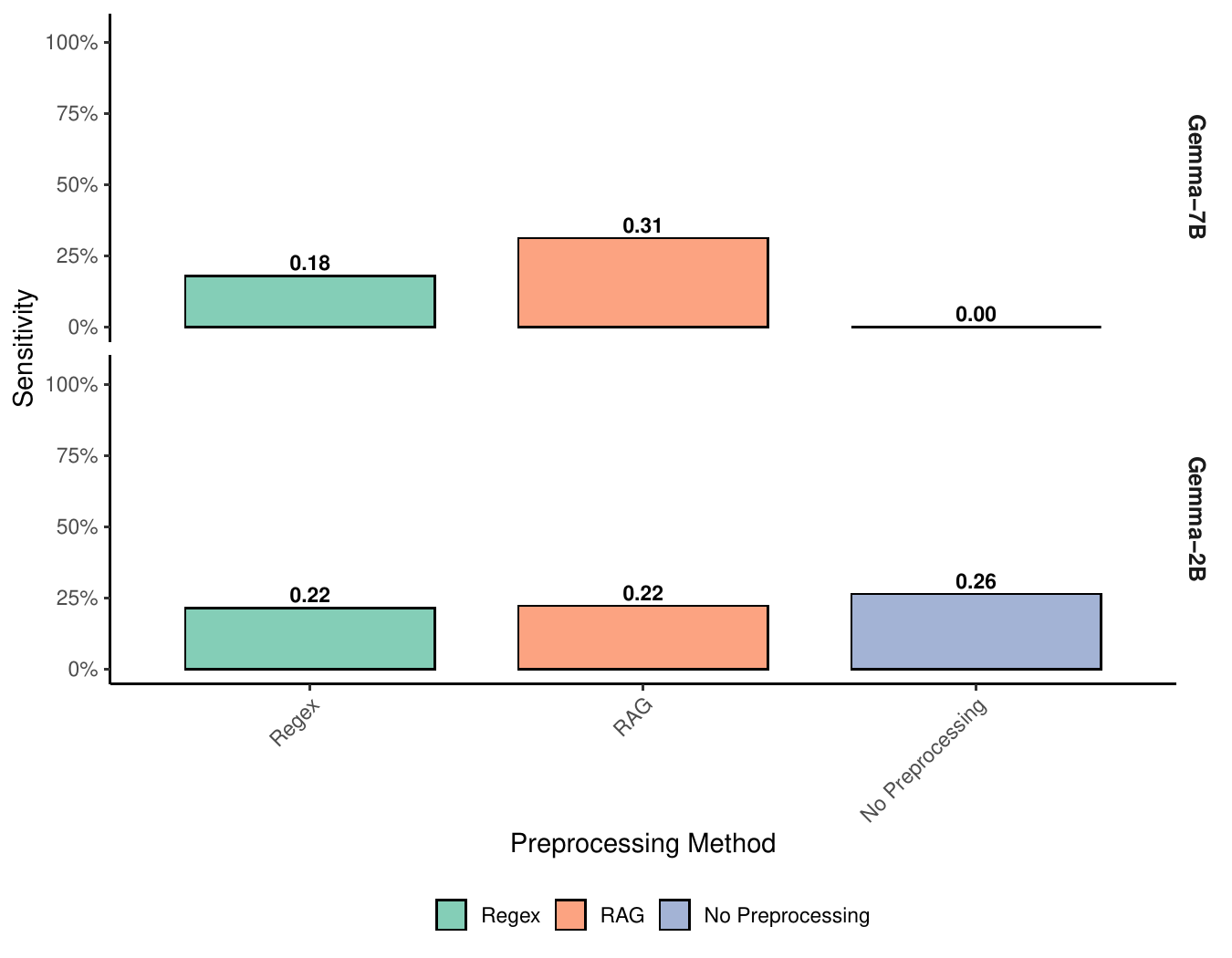}
        \caption{Model sensitivity for metastasis detection at the hospital admission level using fine-tuned Gemma models.}
        \label{fig:hadm_model_sens}
    \end{subfigure}

    \begin{subfigure}[b]{0.49\textwidth}
        \centering
        \includegraphics[width=\textwidth]{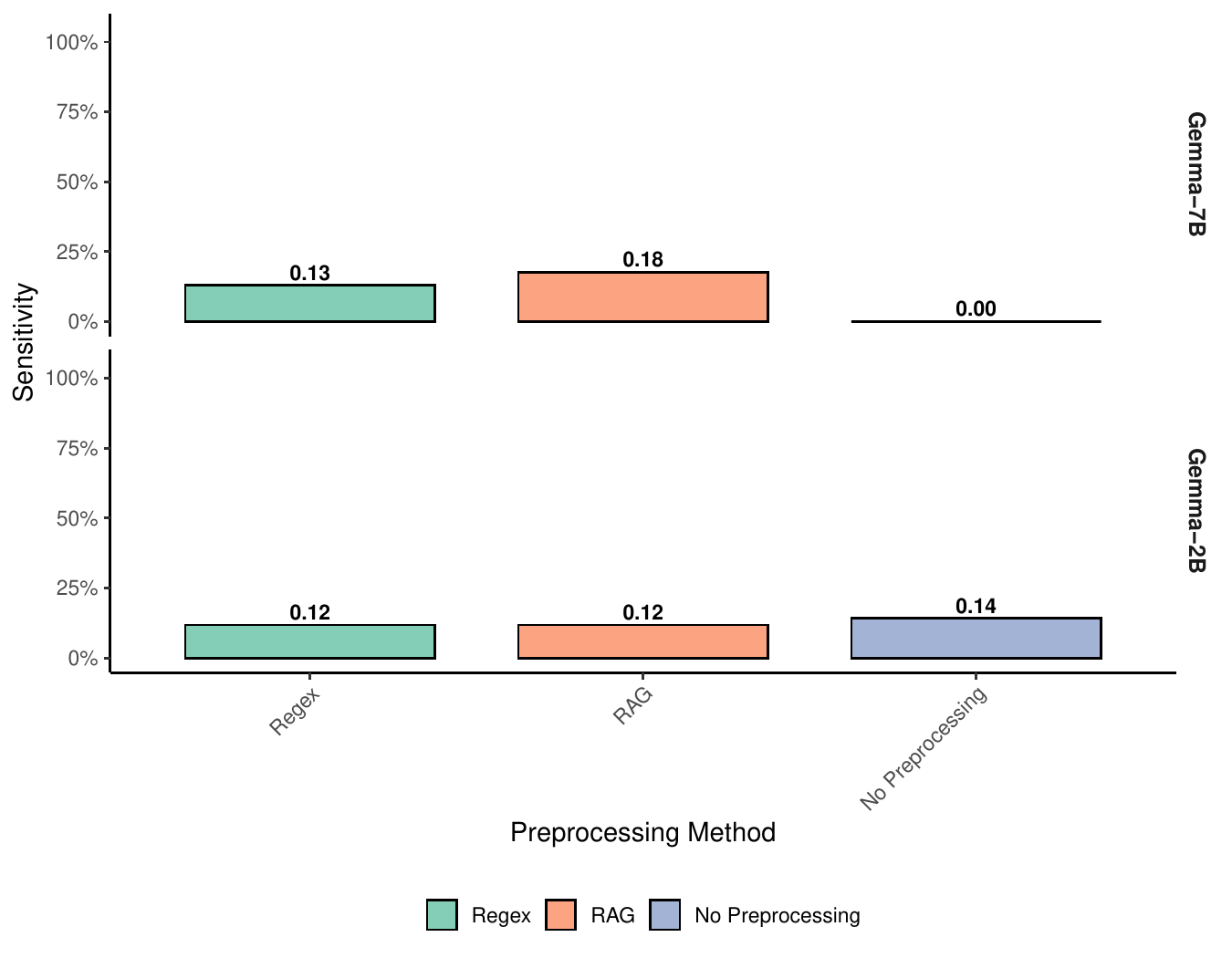}
        \caption{System sensitivity for metastasis detection at the subject level using fine-tuned Gemma models.}
        \label{fig:subject_system_sens}
    \end{subfigure}
    \hfill
    \begin{subfigure}[b]{0.49\textwidth}
        \centering
        \includegraphics[width=\textwidth]{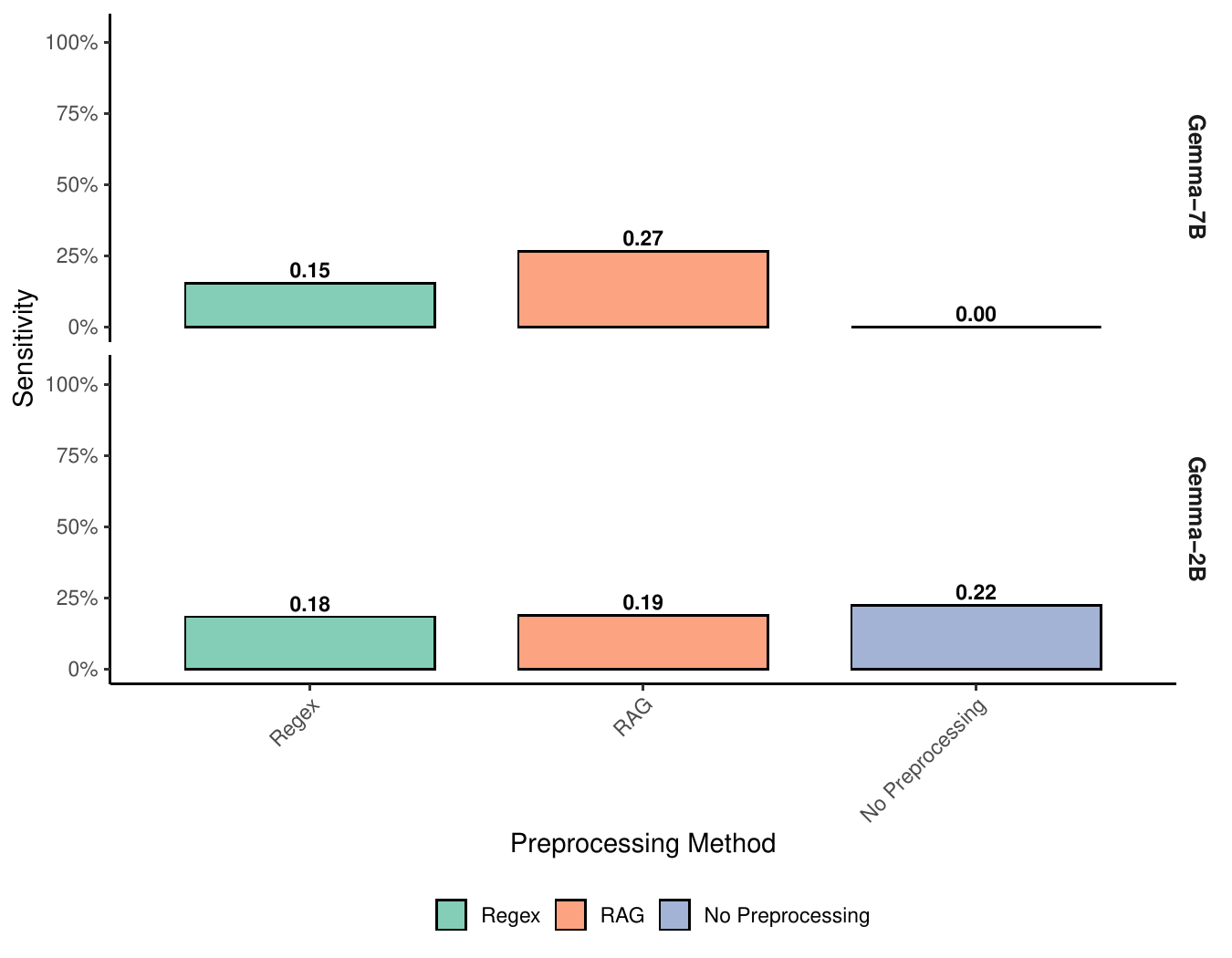}
        \caption{System sensitivity for metastasis detection at the hospital admission level using fine-tuned Gemma models.}
        \label{fig:hadm_system_sens}
    \end{subfigure}

    \caption{Model and system sensitivity for metastasis detection on the MIMIC-IV subset using fine-tuned \texttt{Gemma-2B-it} and \texttt{Gemma-7B-it} models, reported at both the subject level and the hospital admission level.}
    \label{fig:finetune_sensitivity_comparison}
\end{figure*}

\begin{figure*}[ht]
    \centering
    \begin{subfigure}[b]{0.49\textwidth}
        \centering
        \includegraphics[width=\textwidth]{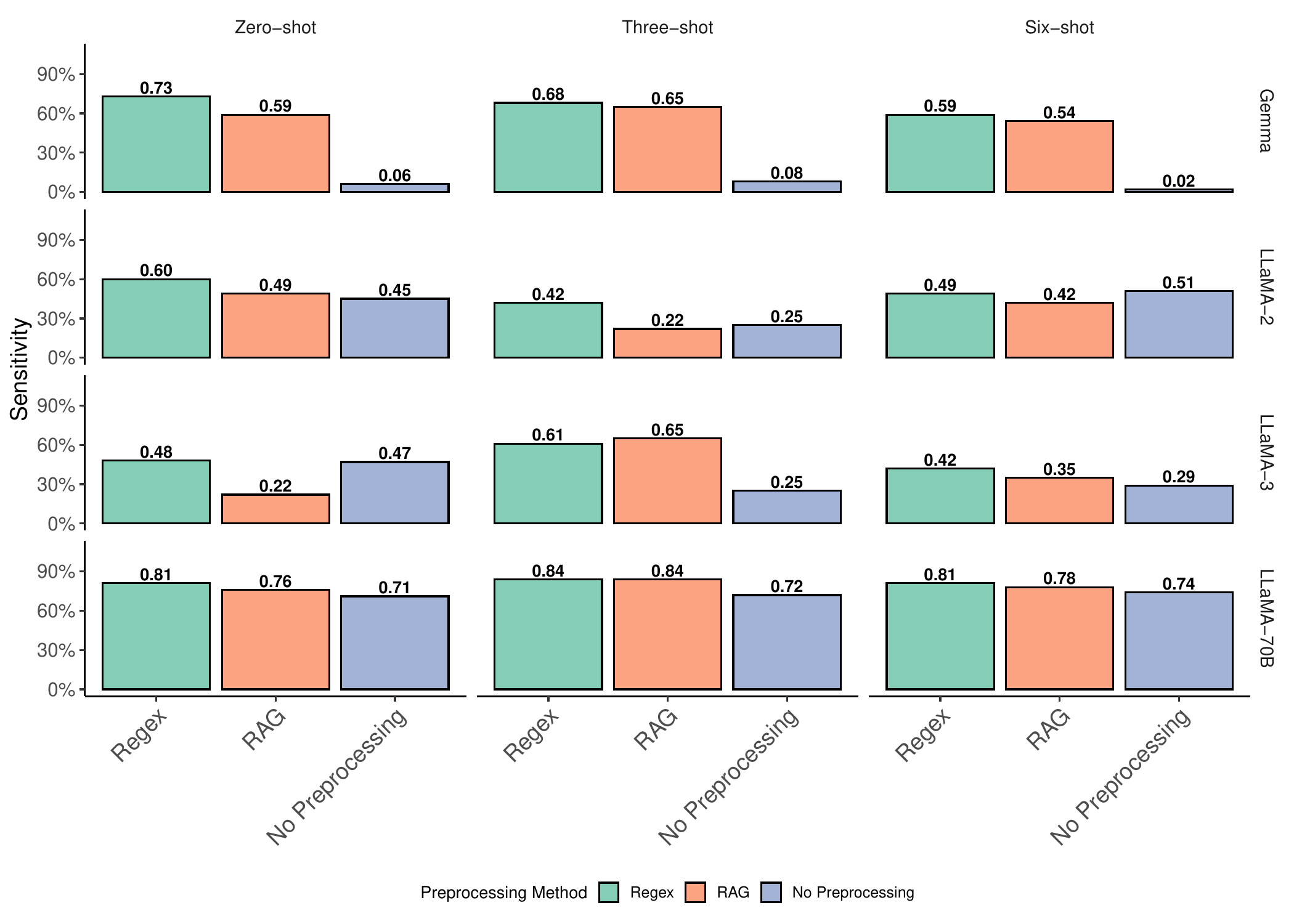}
        \caption{System sensitivity for metastasis detection at the subject level on the MIMIC-IV subset.}
        \label{fig:system_metastasis_subject}
    \end{subfigure}
    \hfill
    \begin{subfigure}[b]{0.49\textwidth}
        \centering
        \includegraphics[width=\textwidth]{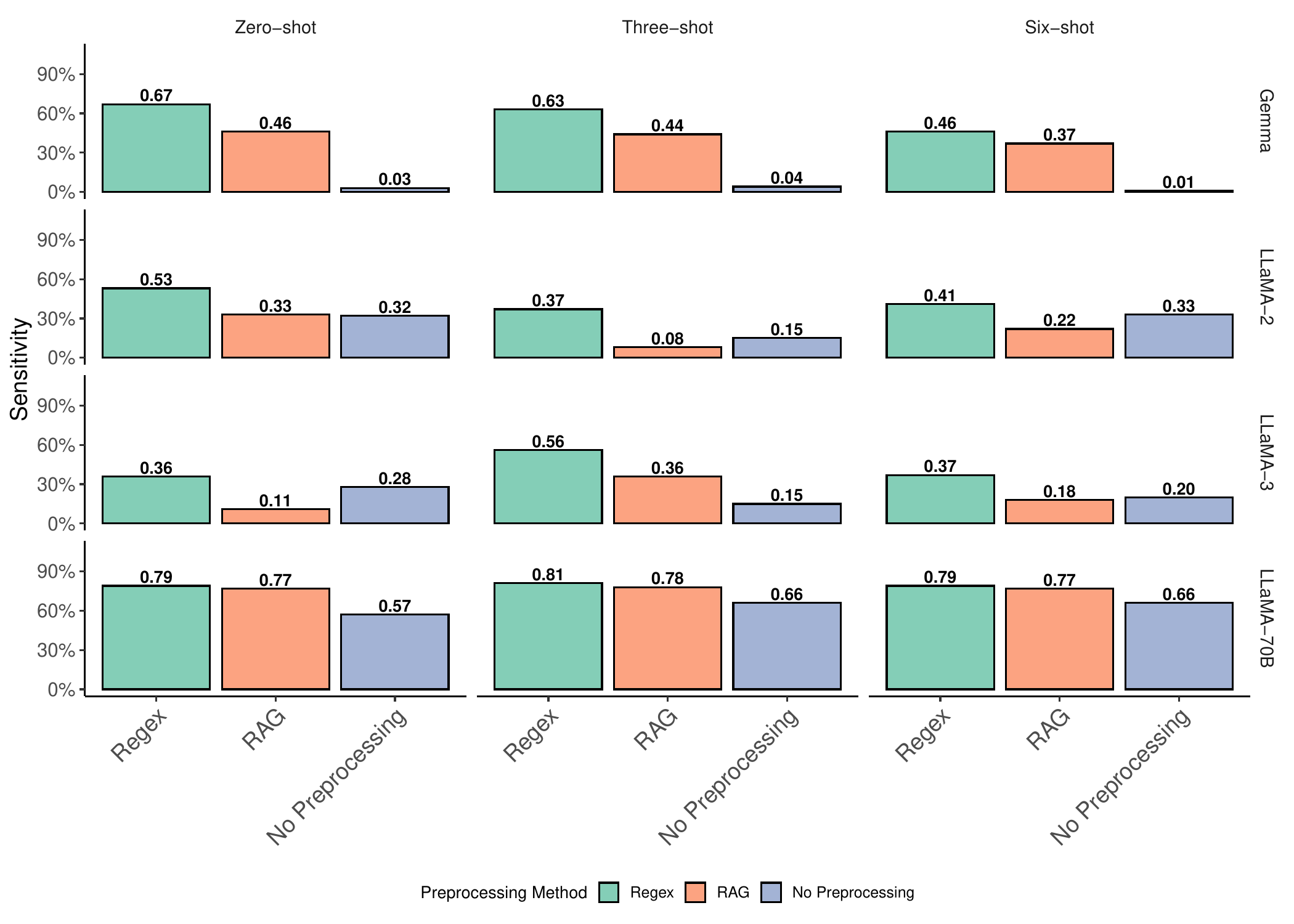}
        \caption{System sensitivity for metastasis detection at the hospital admission level on the MIMIC-IV subset.}
        \label{fig:system_metastasis_hadm}
    \end{subfigure}

    \caption{System sensitivity for metastasis detection on the MIMIC-IV subset at the subject level and the hospital admission level, evaluated across four LLMs under zero-shot, three-shot, and six-shot settings with regex, RAG, and non-preprocessed methods.}
    \label{fig:comparison_system_metastasis}
\end{figure*}

\begin{figure*}[ht]
    \centering
    \begin{subfigure}[b]{0.49\textwidth}
        \centering
        \includegraphics[width=\textwidth]{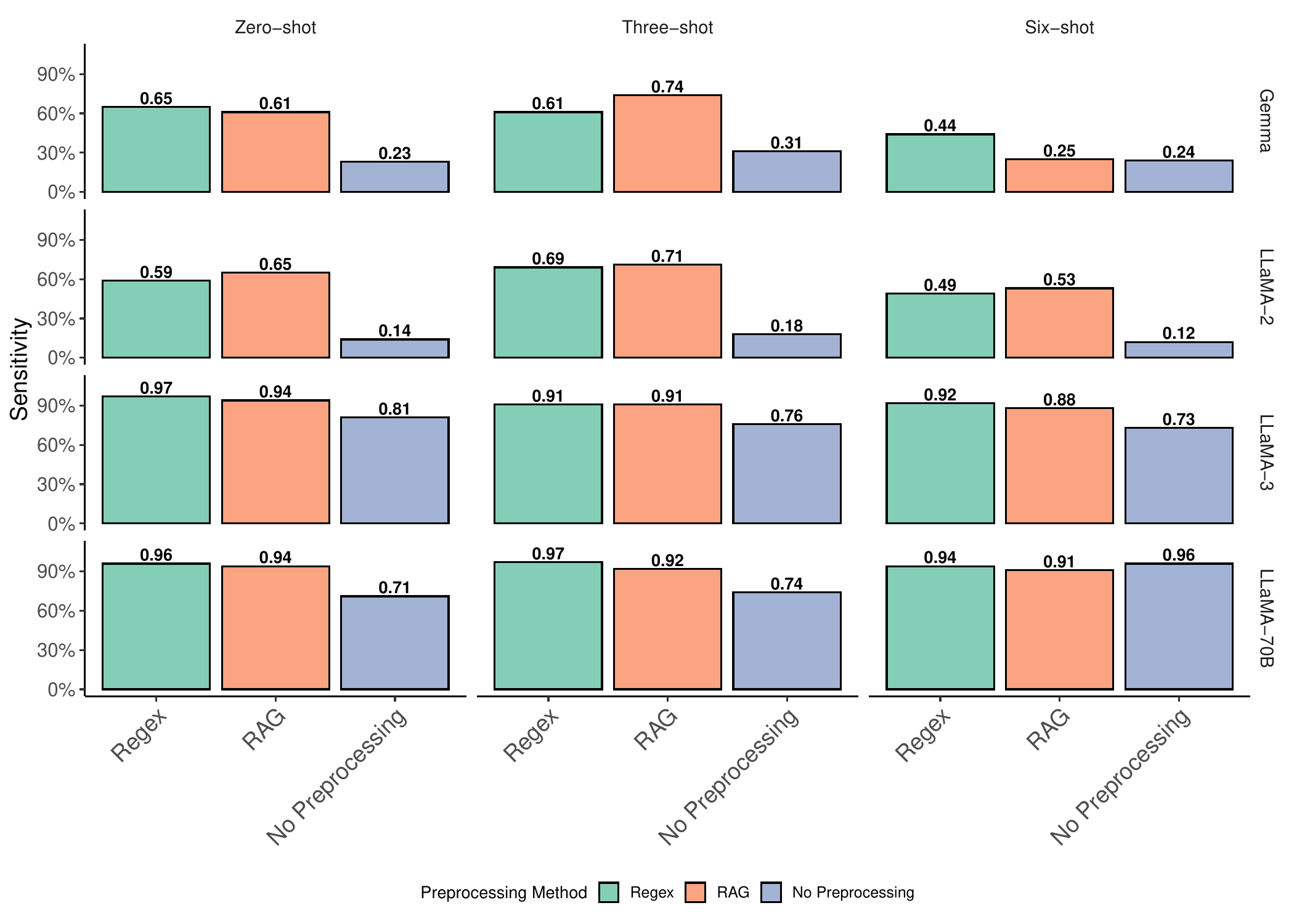}
        \caption{Model sensitivity for insulin use detection at the subject level on the MIMIC-IV subset.}
        \label{fig:insulin_subject}
    \end{subfigure}
    \hfill
    \begin{subfigure}[b]{0.49\textwidth}
        \centering
        \includegraphics[width=\textwidth]{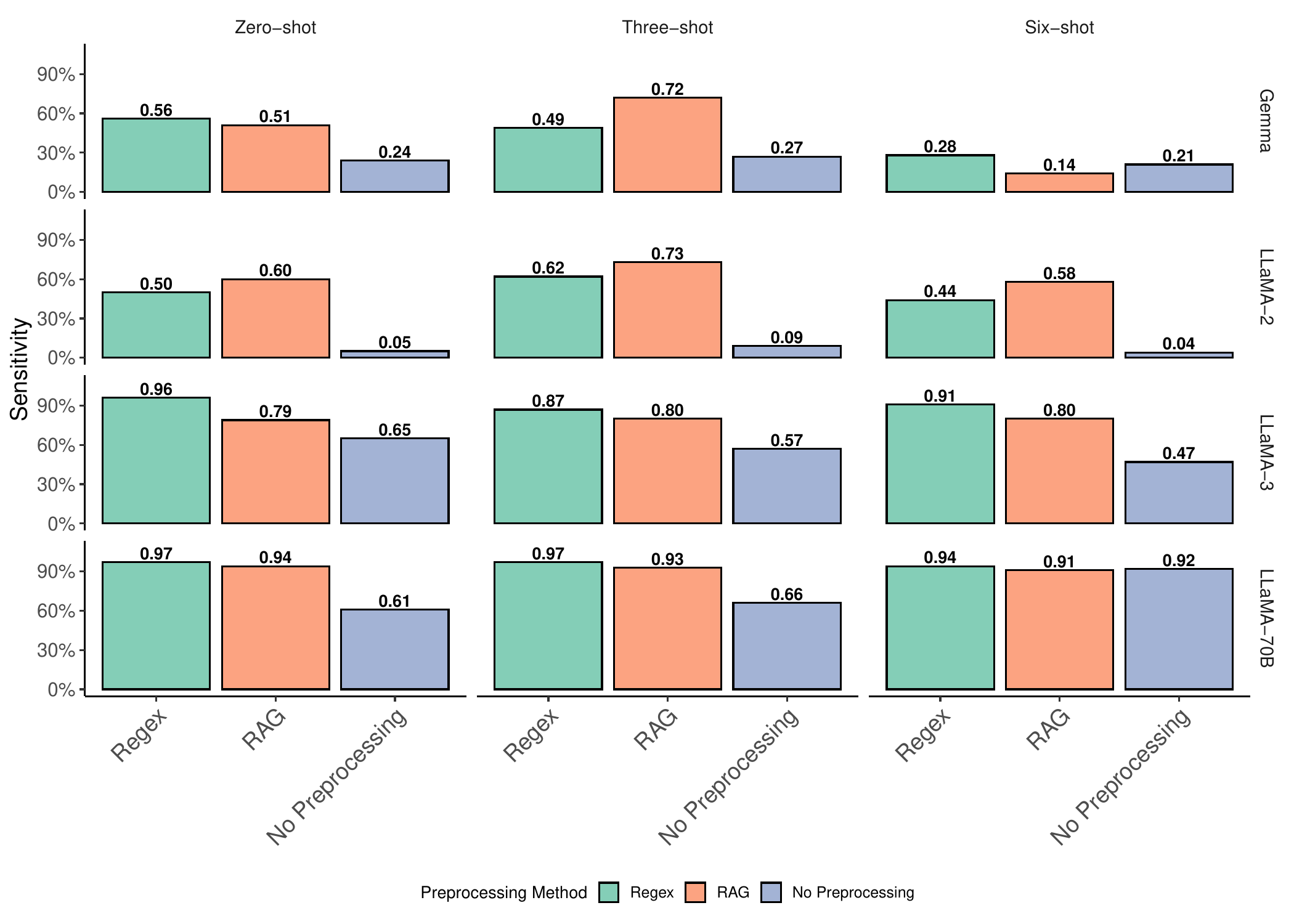}
        \caption{Model sensitivity for insulin use detection at the hospital admission level on the MIMIC-IV subset.}
        \label{fig:insulin_hadm}
    \end{subfigure}

    \caption{Model sensitivity for insulin use detection on the MIMIC-IV subset at the subject level and the hospital admission level, evaluated across four LLMs under zero-shot, three-shot, and six-shot settings with regex, RAG, and non-preprocessed methods.}
    \label{fig:comparison_insulin}
\end{figure*}

\begin{figure*}[ht]
    \centering
    \begin{subfigure}[b]{0.49\textwidth}
        \centering
        \includegraphics[width=\textwidth]{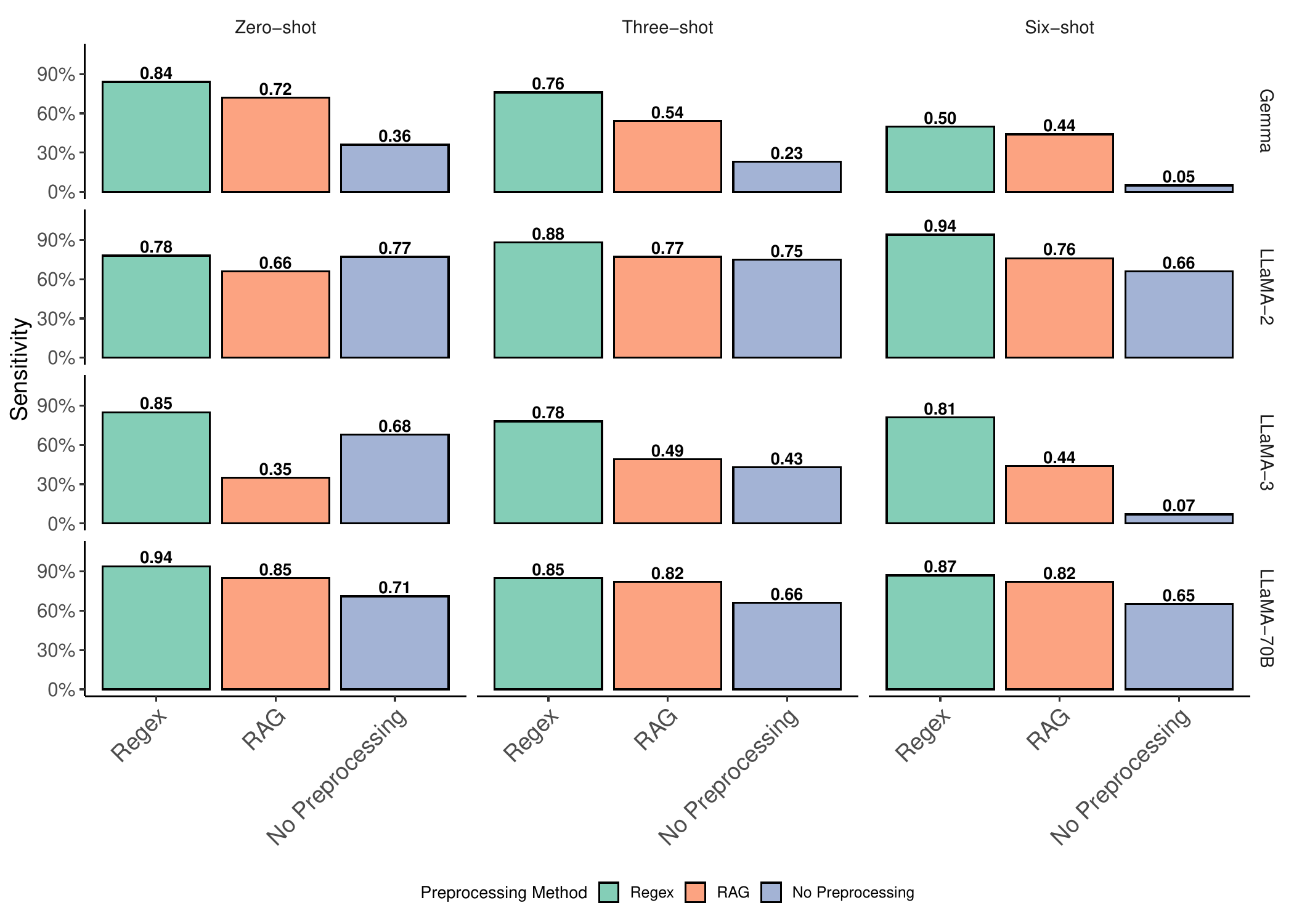}
        \caption{Model sensitivity for hypertension detection at the subject level on the MIMIC-IV subset.}
        \label{fig:hypertension_subject}
    \end{subfigure}
    \hfill
    \begin{subfigure}[b]{0.49\textwidth}
        \centering
        \includegraphics[width=\textwidth]{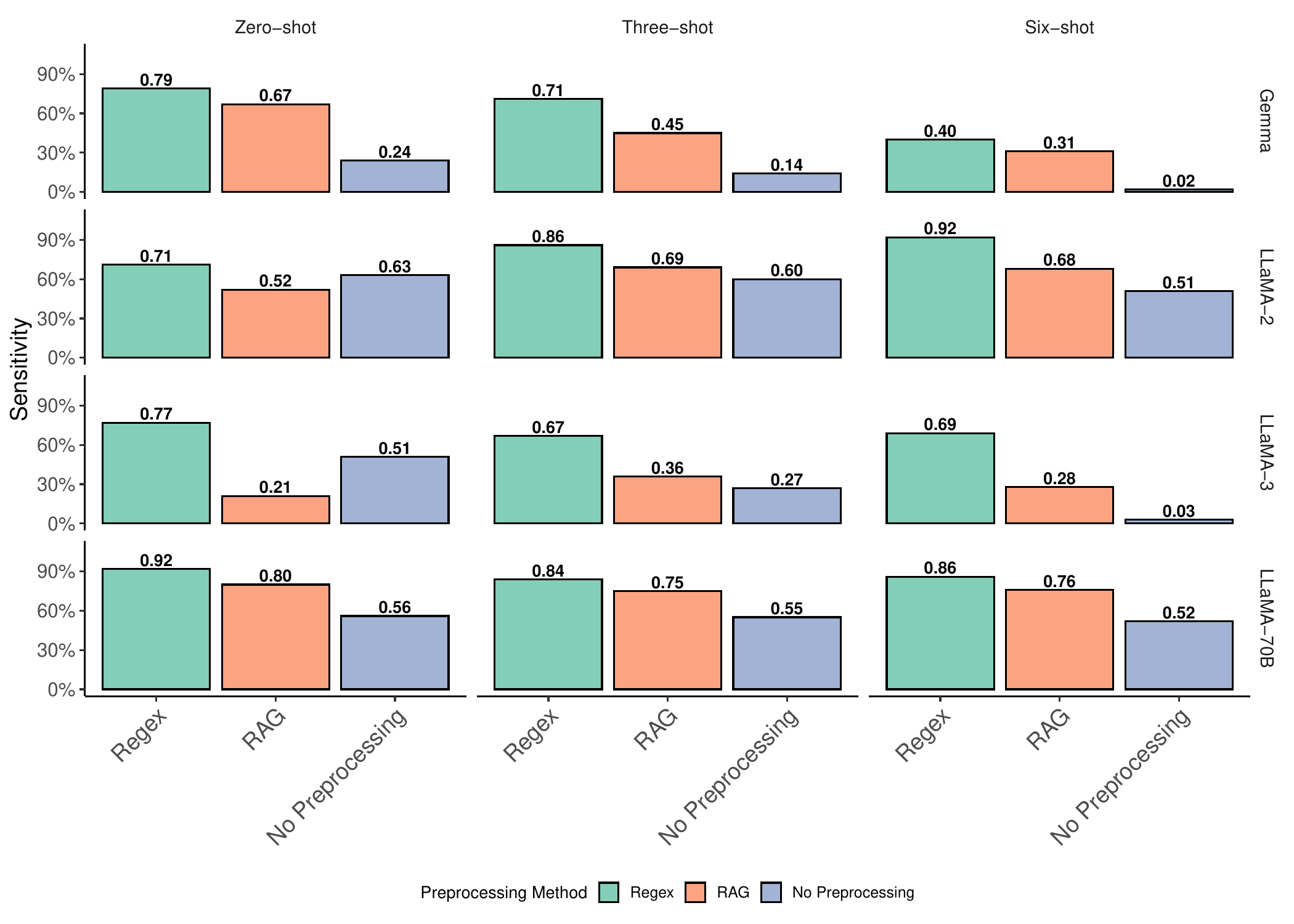}
        \caption{Model sensitivity for hypertension detection at the hospital admission level on the MIMIC-IV subset.}
        \label{fig:hypertension_hadm}
    \end{subfigure}

    \caption{Model sensitivity for hypertension detection on the MIMIC-IV subset at the subject level and the hospital admission level, evaluated across four LLMs under zero-shot, three-shot, and six-shot settings with regex, RAG, and non-preprocessed methods.}
    \label{fig:comparison_hypertension}
\end{figure*}

\begin{figure*}[ht]
    \centering
    \begin{subfigure}[b]{0.49\textwidth}
        \centering
        \includegraphics[width=\textwidth]{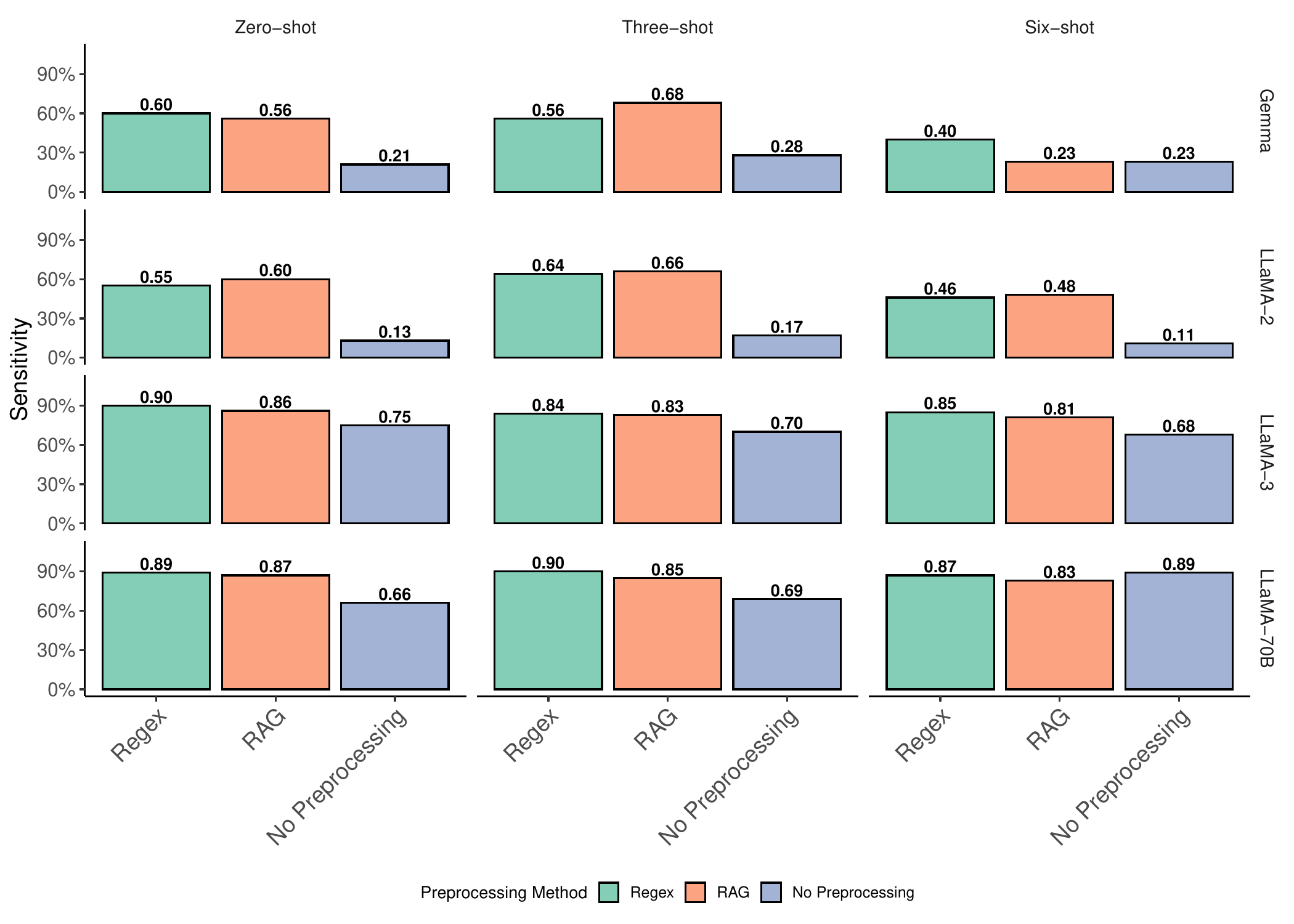}
        \caption{System sensitivity for insulin use detection at the subject level on the MIMIC-IV subset.}
        \label{fig:system_insulin_subject}
    \end{subfigure}
    \hfill
    \begin{subfigure}[b]{0.49\textwidth}
        \centering
        \includegraphics[width=\textwidth]{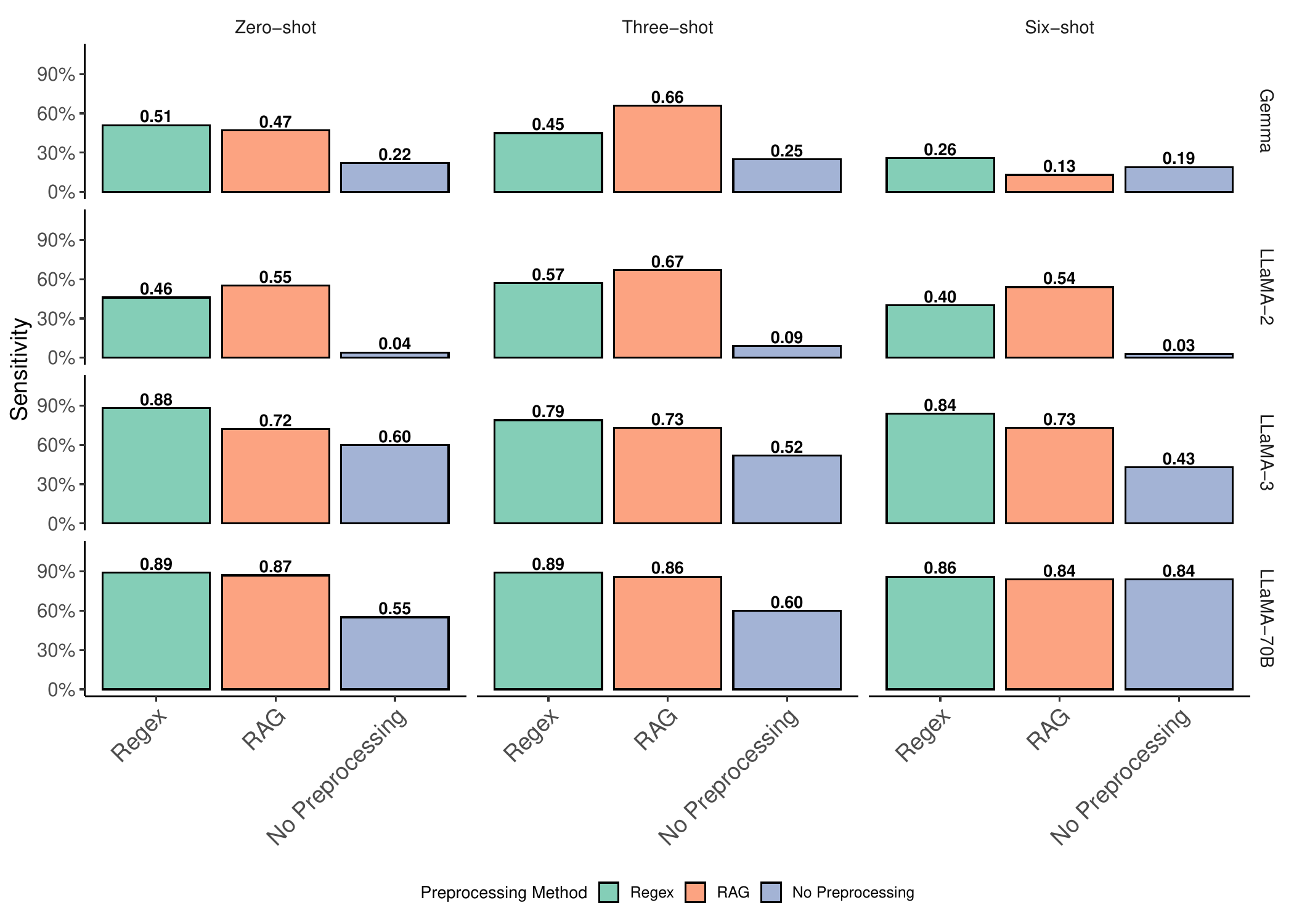}
        \caption{System sensitivity for insulin use detection at the hospital admission level on the MIMIC-IV subset.}
        \label{fig:system_insulin_hadm}
    \end{subfigure}

    \caption{System sensitivity for insulin use detection on the MIMIC-IV subset at the subject level and the hospital admission level, evaluated across four LLMs under zero-shot, three-shot, and six-shot settings with regex, RAG, and non-preprocessed methods.}
    \label{fig:comparison_system_insulin}
\end{figure*}

\begin{figure*}[ht]
    \centering
    \begin{subfigure}[b]{0.49\textwidth}
        \centering
        \includegraphics[width=\textwidth]{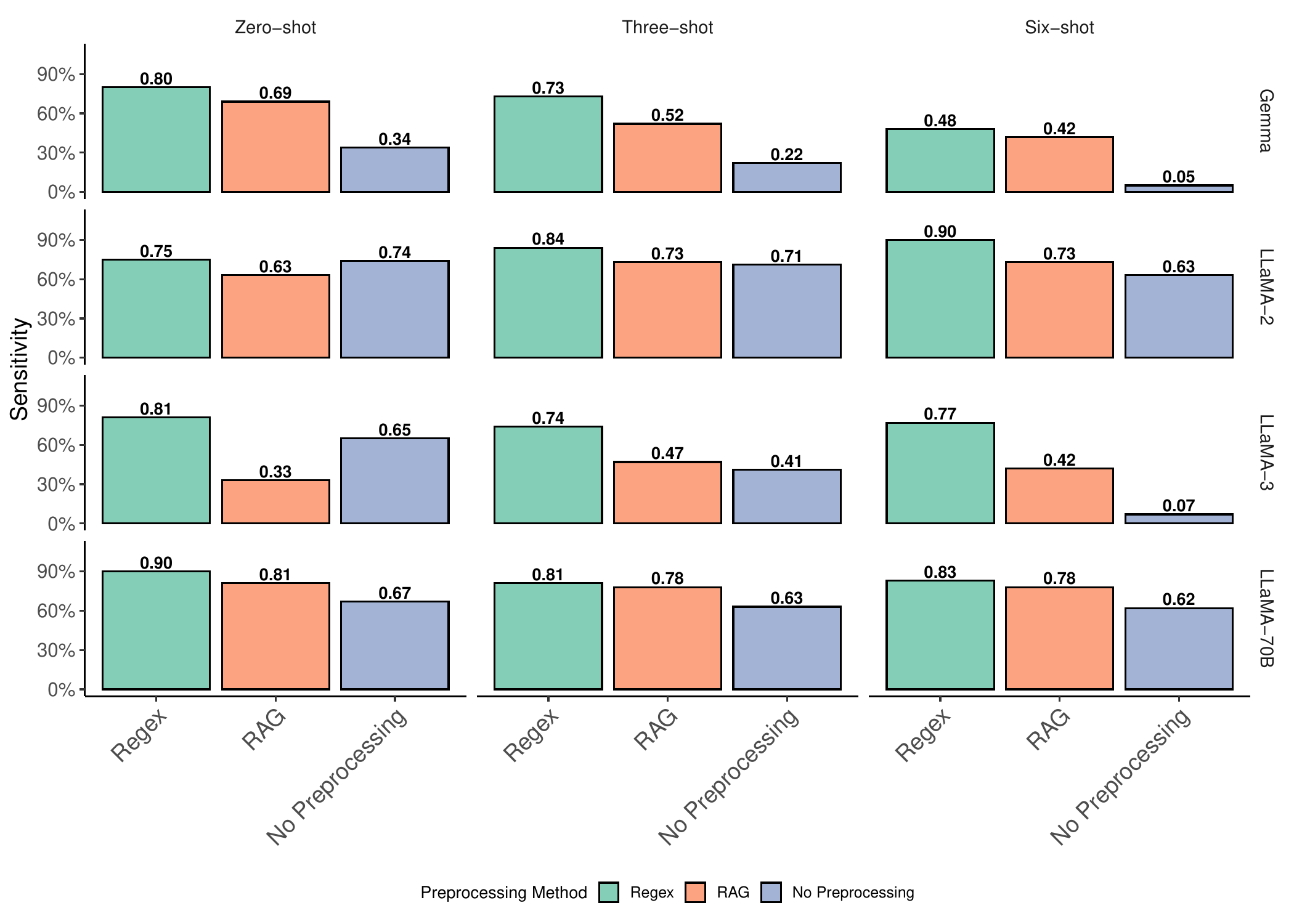}
        \caption{System sensitivity for hypertension detection at the subject level on the MIMIC-IV subset.}
        \label{fig:system_hypertension_subject}
    \end{subfigure}
    \hfill
    \begin{subfigure}[b]{0.49\textwidth}
        \centering
        \includegraphics[width=\textwidth]{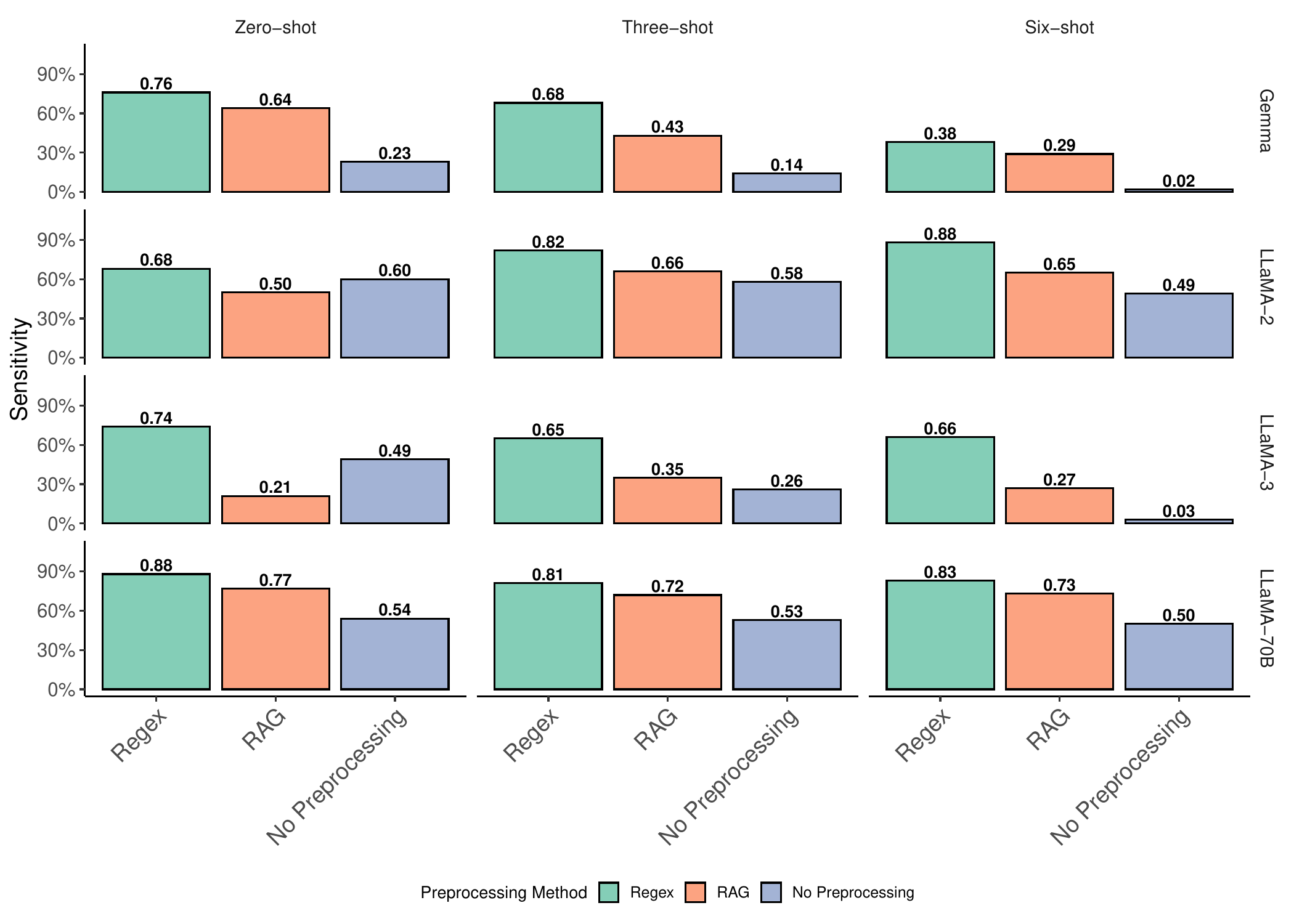}
        \caption{System sensitivity for hypertension detection at the hospital admission level on the MIMIC-IV subset.}
        \label{fig:system_hypertension_hadm}
    \end{subfigure}

    \caption{System sensitivity for hypertension detection on the MIMIC-IV subset at the subject level and the hospital admission level, evaluated across four LLMs under zero-shot, three-shot, and six-shot settings with regex, RAG, and non-preprocessed methods.}
    \label{fig:comparison_system_hypertension}
\end{figure*}

\clearpage

\noindent The following sections provide supplementary details referenced in the main text, including LLM specifications and RAG implementation details (Supplementary Note~\ref{Sec:implementation}), the keyword lists used for regex-based preprocessing (Supplementary Note~\ref{Sec:keywords}), the ICD code definitions adopted as ground-truth labels (Supplementary Note~\ref{Sec:ICD_code}), and the prompt templates used for LLM inference (Supplementary Note~\ref{Sec:prompt}).

\section{LLM Specifications and RAG Implementation Details}
\label{Sec:implementation}

\subsection{Gemma-7B-it}

The primary small LLM used in PrecLLM was \texttt{Gemma-7B-it} \textbf{Version~1} (released February 2024 by Google DeepMind), the instruction-tuned variant with 7 billion parameters. The model weights were obtained via Kaggle Hub (\texttt{google/gemma/pyTorch/7b-it}) and deployed locally on NVIDIA GPUs using the official \texttt{gemma\_pytorch} library in bfloat16 precision. The instruction-tuned variant was chosen over the base model because it was optimized for following structured prompts and producing constrained outputs, which was essential for the three-way classification task. Generation parameters were set as follows: temperature~$= 0.1$, top-$p = 1.0$, top-$k = 100$, and maximum output length~$= 3$ tokens. The low temperature was chosen to produce near-deterministic outputs suitable for classification.

\subsection{LLaMA-Family Models}

Three additional LLaMA-family models were evaluated to assess the generalizability of PrecLLM across different model architectures and scales. All three were loaded via the Hugging Face \texttt{transformers} library (\texttt{AutoModelForCausalLM}) with quantization provided by \texttt{bitsandbytes}, and shared the same core generation settings: \texttt{do\_sample=True}, temperature~$= 0.1$, and \texttt{max\_new\_tokens}~$= 4$.

\begin{itemize}[nosep, leftmargin=1.5em]
    \item \textbf{LLaMA-2-7B-Chat-Med} (\texttt{lianggq/llama-2-7b-chat-med}): A medical-domain fine-tune of LLaMA~2 with 7B parameters. Loaded in 8-bit quantization on a single GPU.
    \item \textbf{Bio-Medical-LLaMA-3-8B} (\texttt{ContactDoctor/Bio-Medical-Llama-3-8B}): A biomedical fine-tune of LLaMA~3 with 8B parameters. Loaded in 8-bit quantization on a single GPU.
    \item \textbf{Meta-Llama-3-70B-Instruct} (\texttt{meta-llama/Meta-Llama-3-70B-Instruct}): The instruction-tuned LLaMA~3 model with 70B parameters, included to evaluate PrecLLM at a larger scale. Loaded in 8-bit quantization with automatic multi-GPU device mapping.

\end{itemize}

\subsection{RAG-Based Preprocessing Setup}

As described in the main text, the RAG-based preprocessing served as an alternative to regex when domain knowledge for keyword curation was limited. The semantic retrieval component was implemented using the \texttt{sentence-transformers} library (Hugging Face) with the \texttt{all-MiniLM-L6-v2} model, a lightweight sentence embedding model based on Microsoft's MiniLM architecture that produces 384-dimensional dense vectors. Nearest-neighbor retrieval was performed using FAISS (Facebook AI Similarity Search) with an \texttt{IndexFlatL2} index, computing L2 (Euclidean) distances between sentence embeddings and the target query embedding.

The key hyperparameters for RAG-based retrieval were:

\begin{itemize}[nosep, leftmargin=1.5em]
    \item \textbf{Top-$k$ sentences}: Number of candidate sentences retrieved per note. Two settings were evaluated: $k = 2$ (denoted RAG2) and $k = 6$ (denoted RAG6).
    \item \textbf{Distance threshold}: $1.5$ (L2 distance). A sentence was retained only if its distance to the query embedding was below this threshold, even when ranked in the top~$k$.
\end{itemize}

For each clinical note, sentences were first segmented using regex-based boundary detection (splitting on double newlines or sentence-ending punctuation), with short fragments ($<5$ characters) merged into adjacent sentences. Each sentence was then encoded into a 384-dimensional embedding, and FAISS retrieved the top-$k$ most similar sentences to the query. Only sentences passing both the top-$k$ and distance-threshold criteria were forwarded to the LLM for classification.

\section{Keyword Lists for Regex-Based Preprocessing}
\label{Sec:keywords}

As described in the main text, the regex-based preprocessing step relies on a curated set of Semantically Similar Terms (SSTs) for each target clinical variable. These keyword lists were constructed through a combination of clinical domain expertise and generative AI assistance (e.g., GPT-4), following three guiding principles: (1) inclusion of canonical medical terms and their morphological variants (e.g., ``metastasis'', ``metastatic'', ``metastasized''); (2) coverage of common clinical abbreviations and shorthand notations (e.g., ``HTN'' for hypertension, ``IDDM'' for insulin-dependent diabetes mellitus); and (3) incorporation of semantically related concepts that frequently co-occur with the target variable in clinical narratives (e.g., ``hematogenous spread'' for metastasis). The keywords were organized into a single regex search command using two complementary matching strategies: flexible patterns for structural variations and comprehensive synonym lists.

\subsection{Metastasis Keywords}

The following keywords capture the broad spectrum of terminology used to describe metastatic disease in clinical notes, ranging from primary diagnostic terms to specific pathological mechanisms:

\begin{itemize}[nosep, leftmargin=1.5em]
    \item \textbf{Metastasis}, \textbf{Metastatic}, \textbf{Metastasize}, \textbf{Metastases}, \textbf{Metastasized}: Primary terms and morphological variants describing the spread of cancer from the original site to other parts of the body.
    \item \textbf{Dissemination}, \textbf{Distant spread}: General terms indicating cancer spread beyond local or regional boundaries.
    \item \textbf{Metachronous}, \textbf{Hematogenous spread}, \textbf{Lymphatic spread}: Terms specifying the temporal pattern or anatomical pathway of metastatic dissemination.
    \item \textbf{Micrometastases}, \textbf{Infiltration}: Terms referring to microscopic tumor deposits and tissue penetration, respectively.
    \item \textbf{Tumor spread}, \textbf{Extra-nodal extension}: Terms describing progressive tumor expansion and extension beyond the lymph node capsule.
\end{itemize}

\subsection{Hypertension Keywords}

Hypertension is documented in clinical notes using a relatively small but highly standardized set of terms and abbreviations:

\begin{itemize}[nosep, leftmargin=1.5em]
    \item \textbf{Hypertension}, \textbf{HTN}: Primary terms denoting elevated blood pressure as a chronic condition.
    \item \textbf{BP elevated}, \textbf{Blood pressure elevated}: Descriptive phrases indicating acute or chronic elevation in blood pressure measurements.
    \item \textbf{Hypertensive}: Adjective form used to describe conditions, crises, or states related to high blood pressure (e.g., hypertensive emergency, hypertensive disorder).
\end{itemize}

\subsection{Insulin Use Keywords}

Keywords for insulin use target both the therapeutic intervention itself and the diagnostic labels that imply insulin dependence:

\begin{itemize}[nosep, leftmargin=1.5em]
    \item \textbf{Insulin}, \textbf{Insulin therapy}: Primary terms indicating use of exogenous insulin for glycemic control.
    \item \textbf{Insulin-dependent}, \textbf{IDDM}: Terms describing insulin-dependent diabetes mellitus (Type~1 diabetes or insulin-requiring Type~2 diabetes).
    \item \textbf{Insulin regimen}, \textbf{Insulin dosing}: Phrases related to insulin treatment protocols and dosage adjustments.
\end{itemize}

\section{ICD Code Definitions for Ground-Truth Labels}
\label{Sec:ICD_code}

As noted in the main text, the International Classification of Diseases (ICD) codes served as the practical reference standard (proxy for ground truth) for evaluating PrecLLM's annotation accuracy. Both ICD-9-CM and ICD-10-CM code sets were required because the datasets span the United States coding transition: the private HNC dataset covers 2014--2022 and MIMIC-IV covers 2008--2019, whereas the mandatory switch from ICD-9-CM to ICD-10-CM occurred on October~1, 2015. For each target variable, codes were selected using prefix-based matching on well-established clinical categories, with selection criteria reviewed by domain clinicians to ensure clinical consistency. Supplementary Tables~\ref{tab:icd_metastasis}--\ref{tab:icd_hypertension} list the complete set of codes used.

\begin{table}[ht]
\centering
\caption{ICD codes used to identify metastatic conditions.}
\label{tab:icd_metastasis}
\small
\begin{tabular}{@{}lll@{}}
\toprule
\textbf{Version} & \textbf{Code} & \textbf{Description} \\
\midrule
ICD-9  & 197   & Secondary malignant neoplasm of respiratory and digestive systems \\
ICD-9  & 198   & Secondary malignant neoplasm of other specified sites \\
ICD-9  & 199   & Malignant neoplasm without specification of site \\
\midrule
ICD-10 & C78   & Secondary malignant neoplasm of respiratory and digestive organs \\
ICD-10 & C79   & Secondary malignant neoplasm of other and unspecified sites \\
ICD-10 & C80   & Malignant neoplasm without specification of site \\
\bottomrule
\end{tabular}
\end{table}

\begin{table}[ht]
\centering
\caption{ICD codes used to identify insulin use.}
\label{tab:icd_insulin}
\small
\begin{tabular}{@{}lll@{}}
\toprule
\textbf{Version} & \textbf{Code} & \textbf{Description} \\
\midrule
ICD-9  & V58.67 & Long-term (current) use of insulin \\
\midrule
ICD-10 & Z79.4  & Long-term (current) use of insulin \\
\bottomrule
\end{tabular}
\end{table}

\begin{table}[ht]
\centering
\caption{ICD codes used to identify hypertension.}
\label{tab:icd_hypertension}
\small
\begin{tabular}{@{}lll@{}}
\toprule
\textbf{Version} & \textbf{Code} & \textbf{Description} \\
\midrule
ICD-9  & 401 & Essential hypertension \\
ICD-9  & 402 & Hypertensive heart disease \\
ICD-9  & 403 & Hypertensive chronic kidney disease \\
ICD-9  & 404 & Hypertensive heart and chronic kidney disease \\
ICD-9  & 405 & Secondary hypertension \\
\midrule
ICD-10 & I10 & Essential (primary) hypertension \\
ICD-10 & I11 & Hypertensive heart disease \\
ICD-10 & I12 & Hypertensive chronic kidney disease \\
ICD-10 & I13 & Hypertensive heart and chronic kidney disease \\
ICD-10 & I15 & Secondary hypertension \\
\bottomrule
\end{tabular}
\end{table}

\section{Prompt Templates for LLM Inference}
\label{Sec:prompt}

This section presents the prompt templates used for the LLM classification step (Step~2 of the PrecLLM pipeline). Each prompt was designed following three principles: (1)~framing the task as a structured three-way classification (``Yes'', ``No'', ``Unknown'') to reduce output variability and facilitate downstream majority voting; (2)~restricting the model to output only a single numerical code, thereby minimizing hallucinated explanations; and (3)~providing explicit instructions to default to ``Unknown'' when the evidence is ambiguous, which helps preserve the integrity of the majority-vote aggregation. The templates below use metastasis as the illustrative target variable; for other variables (insulin use, hypertension), only the variable name and the in-context examples are substituted while the template structure remains identical. In the few-shot setting, three or six representative examples were manually selected from the training pool to cover affirmative, negative, and ambiguous cases, ensuring the model was exposed to the full range of expected outputs.

\subsection{Zero-Shot Prompt Template}

\begin{tcolorbox}[colback=gray!5!white, colframe=gray!75!black, title=Zero-shot learning task]
\parbox{\textwidth}{
Task: Classify the presence of metastasis from patient clinical notes. Respond only with the following numerical codes based on the notes provided:

- (1) Yes: The notes explicitly confirm the patient has metastasis.
- (2) No: The notes explicitly confirm the patient does not have metastasis.
- (3) Unknown: The notes do not contain sufficient information to determine the patient's metastasis status.

Instructions:
1. Do not provide explanations or reasons for your classification.
2. Use only the information in the notes for your classification.
3. If the notes are ambiguous or lack details regarding metastasis, choose ``(3) Unknown''.
4. Only provide a single numerical code as the response.

Examples:

- Example 1: ``The CT scan shows multiple nodules in the liver consistent with metastasis.'' -- (1)

- Example 2: ``Patient has a history of cancer but no evidence of metastatic disease on recent imaging.'' -- (2)

- Example 3: ``Patient presents for cancer follow-up.'' -- (3)

Analyze the following clinical notes to determine if the patient has metastasis:

\textcolor{blue}{\{INSERT CURRENT CLINICAL NOTES HERE\}}
}
\end{tcolorbox}

\subsection{Few-Shot Prompt Template}

In the few-shot setting, representative annotated examples are inserted before the query segment. The number of examples ($n = 3$ or $n = 6$) is varied to evaluate the trade-off between additional context and input length constraints of smaller LLMs.

\begin{tcolorbox}[colback=gray!5!white, colframe=gray!75!black, title=Few-shot learning task]
\parbox{\textwidth}{
Task: Classify the presence of metastasis from patient clinical notes. Respond only with the following numerical codes based on the notes provided:

- (1) Yes: The notes explicitly confirm the patient has metastasis.
- (2) No: The notes explicitly confirm the patient does not have metastasis.
- (3) Unknown: The notes do not contain sufficient information to determine the patient's metastasis status.

Instructions:
1. Do not provide explanations or reasons for your classification.
2. Use only the information in the notes for your classification.
3. If the notes are ambiguous or lack details regarding metastasis, choose ``(3) Unknown''.
4. Only provide a single numerical code as the response.

Examples:

\textcolor{red}{\{INSERT $n$ ANNOTATED EXAMPLES HERE\}}

Analyze the following clinical notes to determine if the patient has metastasis:

\textcolor{blue}{\{INSERT CURRENT CLINICAL NOTES HERE\}}

}
\end{tcolorbox}

\end{document}